\documentclass{ieeeaccess}
\usepackage{cite}
\usepackage{amsmath,amssymb,amsfonts}
\usepackage{algorithmic}
\usepackage{graphicx}
\usepackage{textcomp}
\usepackage{bm}
\usepackage{amsmath}
\usepackage{amssymb}
\usepackage{algorithm}
\usepackage{mathtools}
\usepackage{amsthm}
\usepackage{multirow}
\usepackage[FIGBOTCAP]{subfigure}
\usepackage{hhline}
\usepackage{colortbl}
\usepackage{url}
\usepackage{fancyhdr}
\usepackage{lipsum}
\usepackage{hyperref}
\usepackage[table]{xcolor}
\DeclareMathOperator{\Tr}{Tr}

\DeclareMathOperator{\e}{e}
\usepackage{color, colortbl}
\definecolor{Gray}{gray}{0.9}
\newcolumntype{g}{>{\columncolor{Gray}}c}
\newcommand{\mycc}{\cellcolor{lightgray}}

\begin{document}
\history{Date of publication xxxx 00, 0000, date of current version xxxx 00, 0000.}
\doi{00.0000/ACCESS.0000.DOI}

\title{Kernel Clustering with Sigmoid Regularization for Efficient Segmentation of Sequential Data}


\author{\uppercase{Tung Doan}\authorrefmark{1} and
\uppercase{Atsuhiro Takasu}\authorrefmark{2,3} (Member, IEEE)}

\address[1]{Hanoi University of Science and Technology, No. 1, Dai Co Viet road, Hanoi, Vietnam}
\address[2]{The Graduate University for Advanced Studies, SOKENDAI, Kanagawa 240-0193 Japan}
\address[3]{National Institute of Informatics, Tokyo 101-8430, Japan}



\corresp{Corresponding author: Tung Doan (e-mail: tungdp@soict.hust.edu.vn).}

\begin{abstract}
	The segmentation of sequential data can be formulated as a clustering problem, where the data samples are grouped into non-overlapping clusters with the constraint that all members of each cluster are in a successive order. A popular algorithm for optimally solving this problem is dynamic programming (DP), which has quadratic computation and memory requirements. Given that sequences in practice are too long, this algorithm is not a practical approach. Although many heuristic algorithms have been proposed to approximate the optimal segmentation, they have no guarantee on the quality of their solutions. In this paper, we take a differentiable approach to alleviate the aforementioned issues. First, we introduce a novel sigmoid-based regularization to smoothly approximate the constraints. Combining it with objective of the balanced kernel clustering, we formulate a differentiable model termed \textit{Kernel clustering with sigmoid-based regularization} (KCSR), where the gradient-based algorithm can be exploited to obtain the optimal segmentation. Second, we develop a stochastic variant of the proposed model. By using the stochastic gradient descent algorithm, which has much lower time and space complexities, for optimization, the second model can perform segmentation on overlong data sequences. Finally, for simultaneously segmenting multiple data sequences, we slightly modify the sigmoid-based regularization to further introduce an extended variant of the proposed model. Through extensive experiments on various types of data sequences performances of our models are evaluated and compared with those of the existing methods. The experimental results validate advantages of the proposed models. Our Matlab source code is available on \href{https://github.com/TungDP/Kernel-CLustering-with-Sigmoid-based-Regularization}{github}.
\end{abstract}


\titlepgskip=-15pt

\maketitle

\section{Introduction}

Recently, there has been an increasing interest in developing machine learning and data mining methods for sequential data. This is due to the exponential growing in number of collected data sequences from applications in wide range of fields, including computer vision \cite{Hu20,Zheng21,Zhou22}, speech processing \cite{Harchaoui09,Seichepine14,Sakran17} , finance \cite{Lavielle07,Si13,Hallac19}, bio-informatics \cite{Vert10,Maidstone17}, climatology \cite{Reeves07,Verbesselt10,Jamali15,Heo22} and traffic monitoring \cite{Levy09,Lung12,Song20}. The main problem associated with analysis of these sequences is that they consists of a huge number of data samples. Therefore, it is desirable to summarize the whole sequences by a much smaller number of the data representatives, alleviating burden for the subsequent tasks.

Such compressed and concise summarization can be obtained via sequence segmentation. More specifically, this aims at partitioning the data sequences into several non-overlapping and homogeneous segments of variable durations, in which some characteristics remain approximately constant. It is widely recognized in the literature that the segmentation of sequential data can be considered as a clustering problem. The difference is that all data samples of each cluster, which represents a segment, are constrained to be in a successive order. Thus, in this paper, we focus on clustering-based methods for segmentation of data sequences. 

In practice, sequential data are often composed of nonlinear and complex segments. Therefore, kernel methods are often applied to map data samples into a new feature space before segmenting. Due to the constraint imposed on the data samples in each cluster, traditional algorithms for clustering are inapplicable to the segmentation problem. \cite{Harchaoui07} proposed an optimal algorithm based on dynamic programming (DP) for segmenting data sequence in the features space, which is associated with a pre-specified kernel and mapping functions. In general, DP has quadratic time and memory complexities. It even induces running time of order $O(n^4)$ \footnote{including time for computing the cost matrix in the feature space \cite{Celisse18}}, where $n$ is the length of the sequence, in practice. Therefore, it is intractable to perform segmentation on long data sequence using DP-based algorithms. To alleviate this issue, many attempts have been made to create approximations to the optimal algorithm. Although a considerable amount of the computational costs are reduced, there are still critical drawbacks remained in the approximation algorithms. Taking pruned DP \cite{Celisse18} and greedy algorithm \cite{Truong19} as representatives. These methods sequentially partition the data sequence, returning one segment boundary (\textit{a.k.a}, change point) at each iteration. This strategy offers a reduction in the computational time. However, its expense is that errors might occur at the earlier steps and they would influence on the subsequent iterations, inducing a huge bias in the final results. Massive memory complexity is also a vital drawback of almost kernel-based methods. They need store the kernel matrix, which requires order of $O(n^2)$ space. Therefore, they are prohibited by themselves from handling extensively long data sequences.

In this paper, we take a different approach to alleviate the aforementioned issues. More precisely, we introduce a novel sigmoid-based regularization, which smoothly approximates the constraints of the segmentation problem. It is then integrated with balanced kernel clustering to perform segmentation on sequential data. Our method owns several preferable characteristics. First, because objective of the proposed model is differentiable w.r.t unconstrained and continuous variables we can easily optimize it using gradient descent {GD} algorithm. Different from the existing methods, which are just heuristic approximations of the optimal segmentation algorithm, our model has a guarantee on quality of the solutions as convergence of the GD algorithm was theoretically proved \cite{Nocedal06}. Second, the proposed model offers the applicability of a more efficient optimization algorithm based on stochastic gradient -- the gradient that is estimated from a subsequence (mini-batch), which is randomly sampled from the original data sequence at each iteration. Therefore, the stochastic variant of our model has much lower time and space complexities, making segmentation of extensively long data sequences possible. Finally, the proposed model is flexible. We can easily modify the sigmoid-based regularization to further form a new extended variant that can simultaneously segment multiple data sequences. Through extensive experiments on various types of sequential data, our models are evaluated and compared with baseline methods. The results validate advantages of the proposed models. In summary, contributions of this paper are as follows

\begin{itemize}
	\item Introduction of sigmoid-based regularization that enables kernel clustering to partition sequential data. Objective of the proposed method called \textit{Kernel clustering with sigmoid regularization} (KCSR) is smooth and can be effectively solved using gradient-based algorithm.
	\item Development of a stochastic variant of KCSR to reduce the memory complexity, which is prominent in almost kernel-based methods that prohibits them from handling large-scale datasets.
	\item Extension of KCSR for simultaneously segmentation of multiple data sequences. 
	\item Extensively empirical evaluation of the proposed methods on widely public datasets shows theirs superiorities over the existing methods.
\end{itemize}

The rest of this paper is organized as follows: In Section \ref{sec:related}, we review related works that perform segmentation based on clustering methods. Next, we briefly presents some background for our proposed models in Section \ref{sec:background}, . Section \ref{sec:main} introduces the proposed model KCSR and its stochastic version. This section also describes how to modify the sigmoid-based regularization to form an extension of KCSR that can simultaneously segment multiple data sequences. After illustrating and discussing experimental results in Section \ref{sec:experiments}, we conclude the paper in Section \ref{sec:conclusion}.

\section{Related works} \label{sec:related}

In this paper, we focus on clustering-based methods for nonlinear segmentation of sequential data. Thus, we will review related works in the literature of kernel segmentation, which sometime is referred to as \textit{offline kernel change point detection (CPD)} \cite{Truong20}. Here, the change points indicate the boundaries between the segments. In addition, we also review \textit{temporal clustering} methods. They have recently gained more and more popularity in the computer vision field, where clustering-based algorithms are employed to segment videos of human motions.

\textbf{Offline kernel change point detection}. According to \cite{Truong20}, almost all offline kernel CPD methods attempt to optimize the objective function as defined in ({\ref{eq:obj_seg}}). This is also the objective of the kernel $k$-means clustering. Based on the search scheme for the segment boundaries, existing methods can be divided into local group, which uses sliding window and global group, which bases on dynamic programming.

The local methods \cite{Harchaoui09a,Harchaoui09b,Gretton12,LiShuang15,Li19} slide a window with a large enough width over the data sequence. They then detect, in the window, a single change point, at which the difference between the preceding and succeeding samples is maximal. Although having low computational cost, these methods is sub-optimal as the whole sequence is not considered when detecting the changes. Our approach is more similar to the global methods, which take all data samples into account for change detection. \cite{Harchaoui07,Arlot19} employed dynamic programming (DP) algorithm to optimally obtain the segment boundaries. However, because DP have time complexity of order $O(n^4)$ (including computational time of the cost matrix \cite{Celisse18} in the feature space), it is impractical for handling long data sequences. To reduce the time complexity, \cite{Truong19} proposed a greedy algorithm that sequentially detects change points one at an iteration. \cite{Celisse18} further reduce the space requirement by introducing pruned DP, which combines low-rank approximation of the kernel matrix and binary segmentation algorithm. Our approach is different from these two methods as it searches for all the segment boundaries simultaneously. In addition, quality of its solutions is guaranteed as convergence to optimum of the gradient descent algorithm employed in our model is theoretically proved \cite{Nocedal06}. Both pruned DP and the greedy algorithm are heuristic approximations of the original DP. Since  sequentially detect the changes, errors at the early iterations are propagated and can not be corrected at the subsequent iterations. 

\textbf{Temporal clustering} refers to the factorization of data sequences into a set of non-overlapping segments, each of which belongs to one of $k$ clusters. Maximum margin temporal clustering (MMTC) \cite{Hoai12} and Aligned clustering analysis (ACA) \cite{Zhou13} divide data sequences into a set of non-overlapping short segments. These subsequences are then partitioned into $k$ classes using unsupervised support vector machine \cite{Hoai12} or kernel $k$-means clustering \cite{Zhou13}. Recently, a branch of methods based on subspace clustering has been proposed. These methods often include two steps. First, given a data sequences $\bm{X} = \left[ \bm{x}_1, \hdots, \bm{x}_n \right]$, they learn a new representation (coding matrix) $\bm{Z} = \left[ \bm{z}_1, \hdots, \bm{z}_n \right]$ that characterizes the underlying subspaces structures and sequential (a.k.a. temporal) information of the original data. Second, the normalized cut algorithm (Ncut) \cite{Shi00} is then utilized for segmentation of $\bm{Z}$.  

To preserve the sequential information in the new representation, \cite{Tierney14,Wu16} proposed a linear regularization of the form $\lvert \lvert \bm{ZR} \rvert \rvert_{1,2}$, where $\bm{R} \in \mathbb{R}^{n \times (n-1)}$ is a lower triangular matrix with $-1$ on the diagonal and $1$ on the second diagonal. By minimizing this regularization jointly with the subspace learning objective, the new representation $\bm{z}_j$ and $\bm{z}_{j+1}$ of the two consecutive samples $\bm{x}_j$ and $\bm{x}_{j+1}$, respectively, are forced to be similar. \cite{Hu20} further integrated a weight matrix into the linear regularization to avoid equally constraining on every pair of consecutive samples. Nevertheless, since the regularization is linear, it is ineffective for handling complex data structure. To leverage this issue, \cite{LiSheng15,Liu17sequential} proposed manifold-based regularization that preserves the sequential information for the local neighborhood data samples. This type of regularization is more preferable \cite{Zheng21} as it often outperforms the linear one in most tests \cite{Clopton17}. Our approach also employs regularization to model sequential characteristics of the data. However, the sequential information is both globally and locally preserved in the proposed methods, thanks to the smoothness of the sigmoid functions. In addition, since the temporal regularization makes representation of consecutive samples become similar, boundaries of the segments become difficult to be identified. Our methods, in contrast, approximate the boundaries by midpoints in the summation of sigmoid functions with high steepness. Therefore, our models are expected to obtain better segmentation accuracy.

Both temporal clustering and offline kernel CPD approaches have to store an affinity graph matrix and/or a kernel matrix, which require memory of order $O(n^2)$. This is also a vital reason that inhibits them from handling long data sequence. Stochastic variant of our method has significantly lower space requirement. At each iteration, it approximates the gradient based on a partial kernel matrix, which corresponds to data samples in the current minibatch. Therefore, memory complexity of Stochastic KCSR is only $O(b^2)$, where $b \ll n$ is the minibatch size. Among the existing methods, only pruned DP in \cite{Celisse18} is capable of handling large-scale data because it employs low-rank approximation of the kernel matrix, which only requires space of order $O(r^2)$, where $r \ll n$ is the rank of the approximation. Comparison between performances of Stochastic KCSR and this algorithm on large datasets will be given in Section \ref{sec:experiments}.

\section{Notations and background} \label{sec:background}

\subsection{Notations}

Throughout this paper, we denote vectors and matrices by bold lower-case and bold uppercase letters, respectively. For a particular matrix $\bm{A}$, its $i^{\text{th}}$ column is denoted as $\bm{a}_i$ and its element at position $(j,i)$ is expressed by $a_{j,i}$ or $A_{j,i}$. The transpose matrix of $\bm{A}$ is denoted by $\bm{A}^\top$. If $\bm{A}$ is a square matrix of size $n$ then its trace is expressed as $\Tr(\bm{A}) = \sum_{i=1}^n A_{i,i}$. If $\bm{A} \in \{0,1\}^{k \times n}$ then for any given element $A_{j,i}$ we have $A_{j,i} = 0$ or $A_{j,i} = 1$ ($\bm{A}$ is a binary matrix). By $a \ll b$, we mean that $a$ is very small in comparison with $b$.

\subsection{Kernel segmentation}

The goal of the segmentation task is to partition a data sequence into several non-overlapping and homogeneous segments of variable durations. Let $\bm{X}=[ \bm{x}_1,...,\bm{x}_n ] \in \mathbb{R}^{d \times n}$ denotes the given sequence of length $n$ and dimension $d$. For the number of segments $k$ that is specified in advance, a valid solution of the $k-$segmentation problem can be represented by an sample-to-segment indicator matrix $\bm{G} \in \{0,1\}^{k \times n}$, whose each element is as follows

\begin{figure}[t] 
	\centering
	\includegraphics[width=0.48\textwidth]{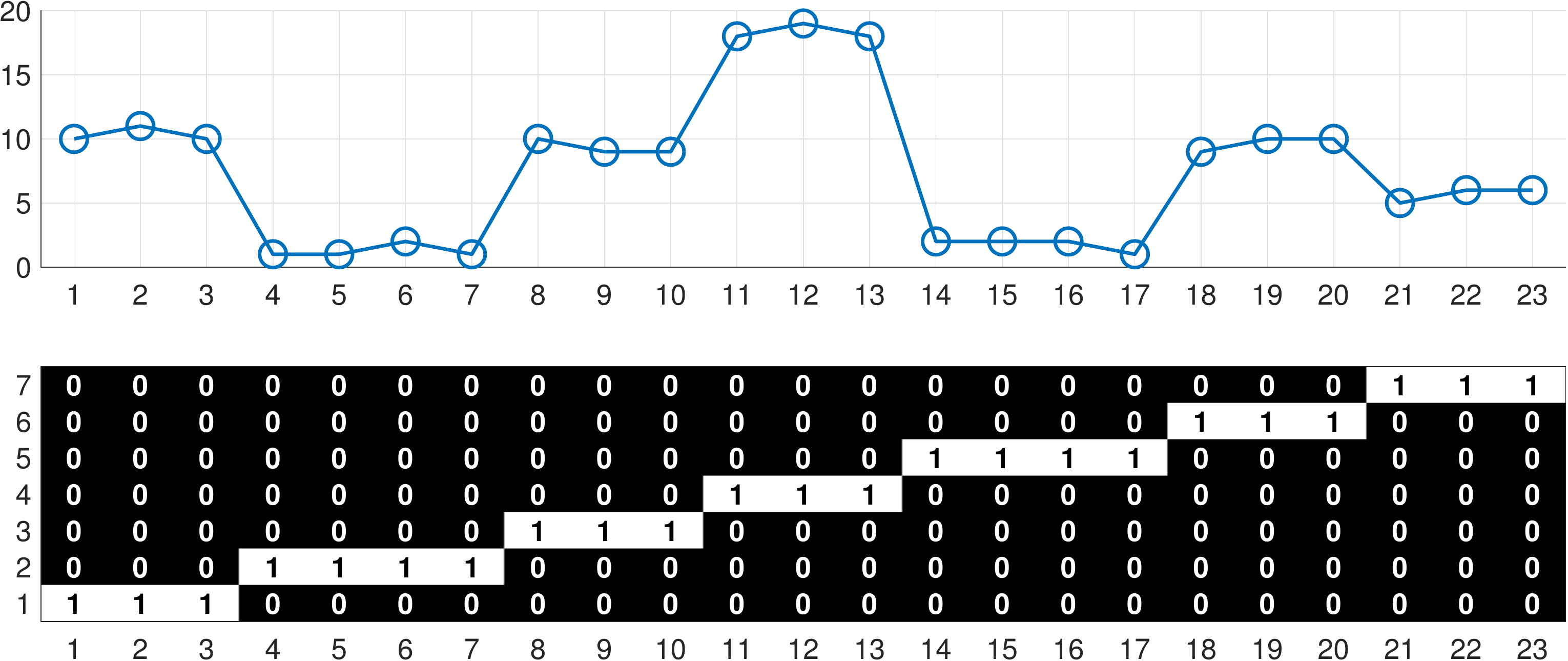}
	\caption{ An example of sequence segmentation: (top) an example sequence of length $23$ and (bottom) the corresponding indicator matrix with number of segments $k = 7$. }
	\label{fig:IndSeg}
\end{figure}

\begin{equation}
	G_{i,j} = \begin{cases}
		1 & \bm{x}_j \in \text{segment } i , \\
		0 & \text{otherwise}.
	\end{cases}
\end{equation}
$\bm{G}$ must satisfy two constraints, including i) \textit{Boundary:} $G_{1,1} = 1$ and $G_{k,n} = 1$ and ii) \textit{Monotonicity:} for any given $G_{i,j} = 1$ then for the next column $G_{i,j+1} = 1$ or $G_{i+1,j+1} = 1$. An example of the indicator matrix is given in Figure \ref{fig:IndSeg}.

To discover segments with complex and nonlinear structures, kernelization is often applied. More specifically, the data sequence $\bm{X}$ is mapped onto some high dimensional space (\textit{a.k.a.} feature space) associated with a pre-specified kernel function $\kappa(\cdot,\cdot): \mathbb{R}^d \times \mathbb{R}^d \rightarrow \mathbb{R}$. The mapping function $\phi(\cdot)$ is implicitly defined by $\phi(\bm{x}_i) = \kappa(\bm{x}_i,\cdot)$, resulting the inner-product $\phi(\bm{x}_i)\phi(\bm{x}_j) = \kappa(\bm{x}_i,\bm{x}_j)$. A common objective for segmentation is to minimize the total summation of the intra-segment variances \cite{Truong20}. Thus, the optimization problem is often formulated as follows
\begin{equation} \label{eq:obj_seg}
	\underset{\bm{G} \, \in \, \mathcal{G}}{\text{argmin}} \, \sum_{j=1}^{k} \sum_{i=1}^{n} \, G_{j,i} \, \lvert \lvert \phi(\bm{x}_i) - \bm{\mu}_j \rvert \rvert_2^2,
\end{equation}
where $\mathcal{G}$ is the set of all valid sample-to-segment indicator matrices and $\bm{\mu}_j$ is the mean of the $j^{\text{th}}$ segment in the feature space. We can observe that the objective of this problem is similar to that of the kernel $k-$means and it is difficult to be minimized because $\bm{G}$ is the discrete variables with combinatorial constraints.

\subsection{Balanced kernel $k-$means}

As mentioned above, kernel segmentation is closely related to kernel $k-$means due to the similarity between their objectives. In fact, this objective can be rewritten in matrix form. More specifically, we can compute the corresponding kernel matrix $\bm{K} \in \mathbb{R}^{n \times n}$, where each element $K_{i,j} = \phi(\bm{x}_i)\phi(\bm{x}_j) = \kappa(\bm{x}_i,\bm{x}_j)$ represents how likely the two samples are assigned to the same class. Let $\bm{G} \in \{0,1\}^{k \times n}$ denotes the associated (unknown) sample-to-class indicator matrix of $\bm{X}$, where $G_{i,j} = 1$ if $\bm{x}_j$ is assigned to the $i^{\text{th}}$ class and zero otherwise. Here, different from the segmentation task, there is no constraint on the indicator matrix $\bm{G}$. Then the objective function of kernel $k-$means \cite{Dhillon04,Torre12,Zass05} can be expressed as follows:
\begin{equation} \label{eq:obj_kkm}
	J_{KKM}(\bm{G}) = \Tr \left( \bm{LK} \right), 
\end{equation}
where $\bm{L} = \bm{I}_n - \bm{G}^\top\left(\bm{G}\bm{G}^\top\right)^{-1}\bm{G}$.

Kernel $k$-means is a strong approach for identifying clusters that are non-linearly separable in the original space. However, similar to its linear counterpart, kernel $k$-means is sensitive to outliers. More specifically, it often outputs unbalanced results that consists of too big and/or too small clusters under presents of anomaly data samples \cite{Zhong03}. To alleviate this issue, recently \cite{Liu17balanced} has proposed a simple regularization on the indicator matrix of the form $\Tr(\bm{G} \bm{1} \bm{1}^\top \bm{G}^\top)$,
where $\bm{1}$ is a vector, whose all elements equal to one. By minimizing this regularization jointly with the clustering objective, we can prevent a too small or too great number of data samples from being partitioned into a cluster. We now can combine (\ref{eq:obj_kkm}) and the regularization to form a new objective of balanced kernel $k$-means
\begin{equation} \label{eq:obj_bkkm}
	J_{BKKM}(\bm{G}) = \Tr \left( \bm{LK} \right) + \lambda \Tr(\bm{G} \bm{1} \bm{1}^\top \bm{G}^\top),
\end{equation}
where $\lambda$ is a positive parameter that controls the balanced regularization.

\section{The proposed method} \label{sec:main}

\subsection{Kernel clustering with sigmoid-based regularization (KCSR) }

\begin{figure}[t]
	\centering
	\subfigure[Clustering task]{ \label{fig:indicator_a}\includegraphics[width=0.46\textwidth]{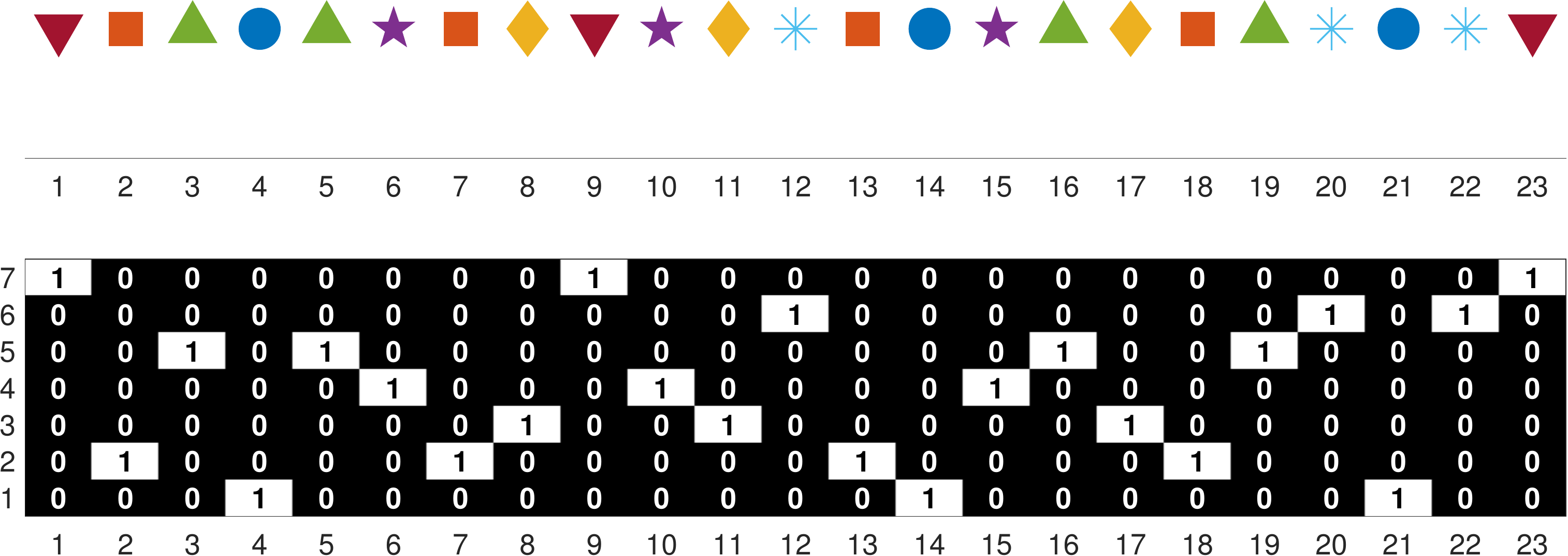} }
	\subfigure[Segmentation task]{ \label{fig:indicator_b}\includegraphics[width=0.46\textwidth]{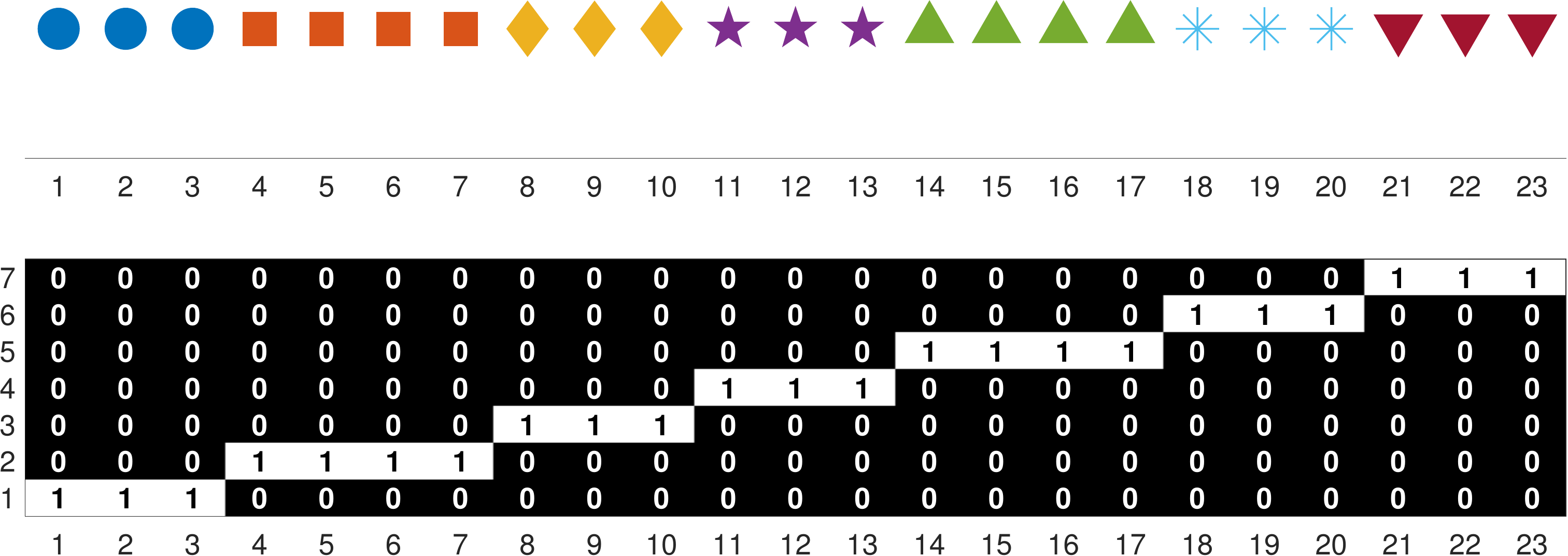} }
	\caption{Toy examples of (a) Clustering task and (b) Segmentation task, where the given data and the corresponding indicator matrix are depicted. Data samples from the same cluster or segment have identical symbol and color. Segmentation is different from clustering in that data samples of the same segment must be in a successive order.}
	\label{fig:indicator}
\end{figure}

Our intuitive idea is to reuse the robust objective of balanced kernel $k-$means (\ref{eq:obj_bkkm}) for segmentation of data sequence $\bm{X} = [\bm{x}_1, \hdots, \bm{x}_n] \in \mathbb{R}^{d \times n}$. However, the challenge is that the sample-to-segment indicator matrix must satisfy two constraints, including \textit{boundary} and \textit{monotonicity}, while the indicator matrix for clustering does not. This difference is illustrated Figure \ref{fig:indicator}. To close this gap and enable the clustering approach to segment data sequences, we introduce a novel regularization that smoothly approximates the two above constraints. The new regularization changes the variables from a discrete to continuous domains. Therefore, our problem can be solved using gradient descent (GD) algorithm. Since, the convergence of GD was already proved \cite{Nocedal06}, quality of the proposed models' solutions is guaranteed.

\begin{figure}[t] 
	\centering
	\includegraphics[width=0.4\textwidth]{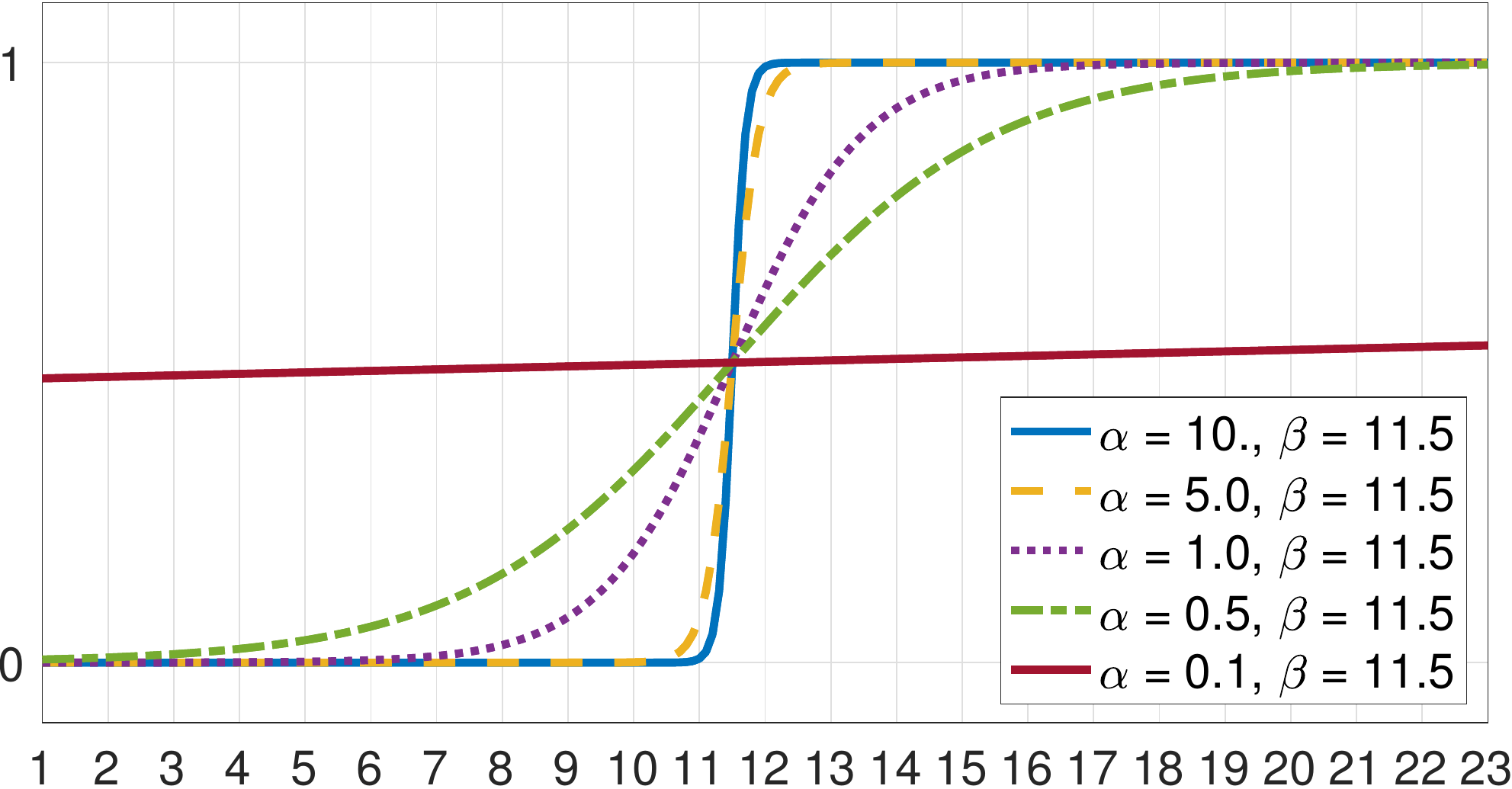}
	\caption{Sigmoid function with different values of the parameter $\alpha$.}
	\label{fig:Sig}
\end{figure}

The proposed regularization is based on the sigmoid function. A basic sigmoid function is defined as
\begin{equation}
	f_{\text{sigmoid}}(x) = \frac{1}{1 + e^{- \alpha (x  - \beta) }},
\end{equation}
where $\beta$ specifies the midpoint and $\alpha$ controls the steepness of the function curve at the midpoint. Figure \ref{fig:Sig} depicts a sigmoid function, where the midpoint $\beta$ is fixed at $11.5$ and the parameter $\alpha$ varies from $0.1$ to $10$.

\begin{figure}[t]
	\centering
	\includegraphics[width=0.45\textwidth]{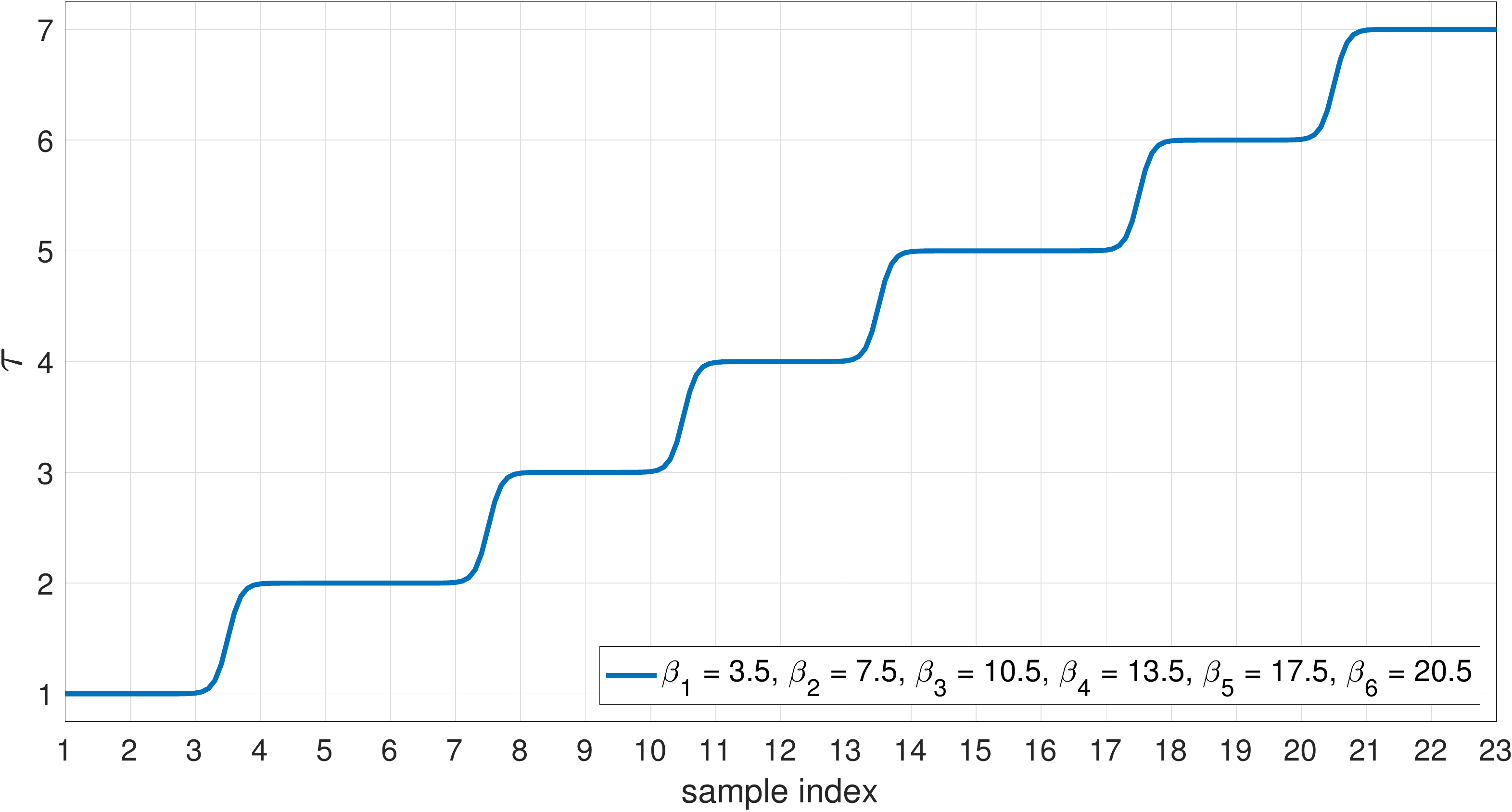}
	\caption{An example of the summation of sigmoid functions with a shared parameter $\alpha = 10$ and $k-1$ different midpoint parameters $\beta_1, \hdots, \beta_{k-1}$, where $k = 7$.}
	\label{fig:Mix}
\end{figure}

We can observer that the higher $\alpha$ is the steeper function curve at the midpoint becomes. In addition, the sigmoid function is monotonic and almost piecewise constant. Therefore, it allows us to roughly partition a sequence into two segments, where the parameter $\beta$ approximates the segment boundary. If we denote $\tau_j \in [1,2]$ (continuously valued) as segment label of sample $\bm{x}_j$, then
\begin{equation}
	\tau_j \approx 1 + f_{\text{sigmoid}}(j,\alpha,\beta).
\end{equation}
For instance, if $\alpha = 10$ and $\beta = 11.5$, then $\tau_j \approx 1$ for $j < 11.5$ and $2$ otherwise. To generalize for cases, where the number of segments $k > 2$, we propose to use a summation of $k-1$ sigmoid functions with different parameters $\beta_i$ for $1 \leq i \leq k-1$.
\begin{equation} \label{eq:mixture}
	\tau_j \approx 1 + \sum_{i=1}^{k-1} f_{\text{sigmoid}}(j,\alpha,\beta_i).
\end{equation} 
Figure \ref{fig:Mix} illustrates an example of a summation of sigmoid functions defined in (\ref{eq:mixture}). Here, the steepness parameter $\alpha$ is shared among the sigmoid functions within the summation. $k-1$ midpoint parameters $\bm{\beta} = [\beta_1, \hdots, \beta_{k-1}]$ approximate the segment boundaries between the $k$ segments. Note that the midpoints must satisfy $1 \leq \beta_1 < \hdots < \beta_{k-1} \leq n$ to guarantee the summation of sigmoid functions monotonically increasing. Thus, we regularize the $\bm{\beta}$ by further introducing $k$ parameters $\gamma_1, \hdots, \gamma_{k}$ such that
\begin{equation} \label{eq:gamma}
	\beta_i = \left(1 - \frac{\sum_{i\prime =1}^{i} e^{\gamma_{i^\prime}} }{\sum_{i^\prime = 1}^{k} e^{\gamma_{i^\prime}} }\right) + n \times \frac{\sum_{i\prime =1}^{i} e^{\gamma_{i^\prime}} }{\sum_{i^\prime = 1}^{k} e^{\gamma_{i^\prime}} }.
\end{equation}
In equation (\ref{eq:gamma}), the ratio $\frac{\sum_{i\prime =1}^{i} e^{\gamma_{i^\prime}}}{\sum_{i^\prime = 1}^{k} e^{\gamma_{i^\prime}}}$ is in the range $[0,1]$. Therefore, $\beta_i$ always satisfies $1 \leq \beta_i \leq n$. In addition, the ratio becomes larger as $i$ increases. This guarantees that $\beta_{i^\prime} < \beta_i$ for $1 \leq i^\prime < i \leq k-1$.

It is notable that the summation of sigmoid functions in Figure \ref{fig:Mix} smoothly approximates the indicator matrix $\bm{G}$ of segmentation example in Figure \ref{fig:indicator_b}. To make the observation more clear, we introduce the following approximation to each element of $\bm{G}$ 
\begin{equation} \label{eq:G2tau}
	G_{i,j} \approx \text{max}\left( 0, 1 - \lvert \tau_j - i \rvert \right).
\end{equation}
This equation map the segment label $\tau_j$ from the range $[1,k]$ to the range $[0,1]$ for approximating the sample-to-segment indicator matrix.

We now can formulate an optimization problem that combines objective of the balanced kernel clustering with sigmoid-based regularization for segmentation. Let $\bm{K} \in \mathbb{R}^{n \times n}$ be the kernel matrix of the data sequence $\bm{X}$ then our kernel-based segmentation optimization problem is
\begin{align} \label{eq:obj_kcsr}
	\underset{\gamma_1, \hdots, \gamma_{k}}{\text{argmin }} & \qquad \Tr \left( \bm{LK} \right) + \lambda \Tr(\bm{G} \bm{1} \bm{1}^\top \bm{G}^\top) \\
	\text{ s.t.} \quad & \; \bm{L} = \bm{I}_n - \bm{G}^\top\left(\bm{G}\bm{G}^\top\right)^{-1}\bm{G}, \nonumber \\
	& \; G_{i,j} = \text{max}\left( 0, 1 - \lvert \tau_j - i \rvert \right) \quad \forall i,j, \nonumber \\
	& \; \tau_j  = 1 + \sum_{i=1}^{k-1} f_{\text{sigmoid}}(j,\alpha,\beta_i) \quad \forall j,  \nonumber \\
	& \; \beta_i = \left(1 - \frac{\sum_{i\prime =1}^{i} e^{\gamma_{i^\prime}} }{\sum_{i^\prime = 1}^{k} e^{\gamma_{i^\prime}} }\right) + n \times \frac{\sum_{i\prime =1}^{i} e^{\gamma_{i^\prime}} }{\sum_{i^\prime = 1}^{k} e^{\gamma_{i^\prime}} } \quad \forall i.  \nonumber
\end{align}
Since $\bm{\gamma} = [\gamma_1, \hdots, \gamma_{k}]$ are unconstrained and continuous parameters, we can optimize objective function in (\ref{eq:obj_kcsr}) using the gradient descent algorithm. Let $J(\bm{\gamma})$ denotes the objective function in (\ref{eq:obj_kcsr}), then the gradient w.r.t parameters $\bm{\gamma}$ can be computed using chain rule.
\begin{equation} \label{eq:grad_kcsr}
	\nabla \bm{\gamma} = \frac{\partial J(\bm{\gamma})}{\partial \bm{\gamma}} = \frac{\partial J(\bm{\gamma})}{\partial \bm{G}} \times \frac{\partial \bm{G}}{\partial \bm{\tau}} \times \frac{\partial \bm{\tau}}{\partial \bm{\beta}} \times \frac{\partial \bm{\beta}}{\partial \bm{\gamma}},
\end{equation}
where $\bm{\tau} = [\tau_1, \hdots, \tau_n]$. More details on derivation of the gradient w.r.t $\bm{\gamma}$ is given in Appendix \ref{app:grad}. We call the proposed model \textit{Kernel clustering with sigmoid regularization} (KCSR) and its optimization algorithm is given in Algorithm \ref{alg:gradient_ascent}.

\begin{algorithm}[t]
	\caption{: Gradient descent algorithm for KCSR}
	\label{alg:gradient_ascent}
	\begin{algorithmic}[1]
		\REQUIRE Kernel matrix $\bm{K}$, number of segments $k$, steepness parameter $\alpha$, tolerance $\epsilon$.
		\ENSURE Optimal parameters $\bm{\gamma}^{\ast} = [\gamma^{\ast}_1, \hdots, \gamma^{\ast}_k]^\top$.
		\REPEAT 
		\STATE compute gradient $\nabla \bm{\gamma} = \frac{\partial J}{\partial \bm{\gamma}}$;
		\STATE compute stepsize $\eta$ using Armijo-Goldstein line search \cite{Armijo66,Nocedal06};
		\STATE update $\bm{\gamma}_{(t+1)} = \bm{\gamma}_{(t)} - \eta\nabla \bm{\gamma}_{(t)}$;
		\UNTIL{ $\lvert J(\bm{\gamma}_{(t+1)}) - J(\bm{\gamma}_{(t)}) \rvert \leq \epsilon$ }
	\end{algorithmic}	
\end{algorithm}

\subsection{Stochastic KCSR}

\begin{table*}[t]
	\begin{center}
		\resizebox{\textwidth}{!}{
			\begin{tabular}{|c|c|c|c|c|c|c|c|}
				\hline
				Method & SSC & TSC & ACA & AKS & GKS & \textbf{KCSR} & \textbf{SKCSR} \\
				\hline \hline
				Time & $O(n^2dt + n^2)$ & $O(n^2dt + n^2)$ & $0(n^2n_{\text{max}}t)$ & $O(r^2n + r\log(k)n)$ & $O(kn + n^2)$ & $O(n^2kt + n^2)$ & $O(nbkt + b^2) $ \\
				\hline
				Space & $O(n^2)$ & $O(n^2)$ & $O(n^2n_{\text{max}})$ & $O((k +r)n)$ & $O(n^2)$ & $O(n^2)$ & $O(b^2)$ \\
				\hline
			\end{tabular}
		}
	\end{center}
	\caption{Time and space complexities of different segmentation methods. Here, $n$ denotes length of the data sequence and $k$ is the number of segments. $n_{\text{max}}$ denotes the maximum length of divided subsequences in ACA, $d$ is dimension of the new representation $\bm{Z}$ in OSC and TSC. $t$ denotes number of total iterations. The rank of the approximation of the kernel matrix in AKS is denoted by $r$ and $b$ is the mini-batch size in SKCSR. Note that $b \ll n$.}
	\label{tab:complexities}
\end{table*}

Kernel segmentation allow us to capture nonlinear structure in the data. However, this advantage is achieved at the expense of much higher complexities in both terms of computational time and storage requirement. More specifically, given a sequence of $n$ data samples, existing kernel-based methods compute the kernel matrix $\bm{K}$, whose both time and memory complexities are of order $O(n^2)$. Note that this is also true for temporal clustering methods, where the affinity graph matrix of size $O(n^2)$ is computed and stored while performing Ncut algorithm. When $n$ is large, these methods become computationally difficult. For example, average length of the acceleration data for activity recognition in the experimental section is about $125K$. The corresponding kernel matrix $\bm{K}$ requires up to approximately  $116.4$ GB for storage, which is definitely out of memory for a regular computer.

Our method is also based on the kernel matrix. Especially, at each iteration, our method computes the gradient using the kernel matrix, which makes it very slow and even impossible due to the large memory requirement for handling long data sequences. Fortunately, since objective function of KCSR is differentiable, we can reduce the complexities by using the stochastic gradient descent (SGD) \cite{Robbins51,Bottou98,Spall03}. SGD estimates the gradient from a randomly sampled subsequence\footnote{By sub-sequence, we mean that order and indexes of samples in the original sequence are preserved in the randomly sampled mini-batch.} (a mini-batch), which consists of a much smaller number of samples, from the original sequence. Let $b \ll n$ denotes length of the randomly sampled subsequence $\bm{X}_{(t)}$, where $t$ expresses the iteration index. Then, the stochastic gradient is estimated as follows
\begin{equation} \label{eq:sto_grad_kcsr}
	\nabla \bm{\gamma} = \frac{\partial J(\bm{\gamma})}{\partial \bm{G}_{(t)}} \times \frac{\partial \bm{G}_{(t)}}{\partial \bm{\tau}} \times \frac{\partial \bm{\tau}}{\partial \bm{\beta}} \times \frac{\partial \bm{\beta}}{\partial \bm{\gamma}}.
\end{equation}
In equation (\ref{eq:sto_grad_kcsr}), $\frac{\partial J(\bm{\gamma})}{\partial \bm{G}_{(t)}}$ is only associated with a partial kernel matrix $\bm{K}_{(t)} \in \mathbb{R}^{b \times b}$, which corresponds to the samples in $\bm{X}_{(t)}$. Therefore, it is much more efficient in terms of both running time and memory consumption than computing the full-batch gradient as in equation (\ref{eq:grad_kcsr}). Details of the algorithm is given in Algorithm \ref{alg:stochastic_gradient_descent} and complexity comparison between the proposed methods and several baselines are given in Table \ref{tab:complexities}. We note that convergences of both gradient with step size found by Armijo-Goldstein line search \cite{Armijo66,Nocedal06} and stochastic gradient descent algorithms with vanishing step size are theoretically proven. In fact, it is well-known \cite{Allen-Zhu17, Schmidt17} that gradient descent (GD) after $T$ iterations can find a solution with error $O(T^{-1})$ and stochastic gradient descent (SGD) after $T$ iterations can find a solution with error $O(T^{-0.5})$. Thus, both KCSR and SKCSR can obtain good solutions for problem defined in (\ref{eq:obj_kcsr}) with enough loops.

\begin{algorithm}[t]
	\caption{: Stochastic gradient descent algorithm for KCSR}
	\label{alg:stochastic_gradient_descent}
	\begin{algorithmic}[1]
		\REQUIRE Data sequence $\bm{X}$, number of segments $k$, steepness parameter $\alpha$, number of iterations $T$, minibatch size $b$, initial learning rate $\eta_0$, momentum $\mu \in [0,1)$, weight decay $\rho \in (0,1]$.
		\ENSURE Optimal parameters $\bm{\gamma}^{\ast} = [\gamma^{\ast}_1, \hdots, \gamma^{\ast}_k]^\top$.
		\FOR{$t = 1, \hdots, T$} 
		\STATE $\eta = \eta_0 \times \rho^t$;
		\STATE randomly sample a sub-sequence $\bm{X}_{(t)}$ of length $b$;
		\STATE compute the partial kernel matrix $\bm{K}_{(t)}$;
		\STATE compute the stochastic gradient $\nabla \bm{\gamma} = \frac{\partial J}{\partial \bm{\gamma}}$ based on $\bm{K}_{(t)}$ and original indexes of samples in $\bm{X}_{(t)}$;
		\STATE $\Delta {\gamma}_{(t)} = \eta\nabla \bm{\gamma} - \mu \Delta {\gamma}_{(t - 1)}$;
		\STATE $\bm{\gamma}_{(t)} = \bm{\gamma}_{(t-1)} + \Delta {\gamma}_{(t)}$;
		\ENDFOR
	\end{algorithmic}	
\end{algorithm}

\subsection{Multiple KCSR}

In practice, at some particular circumstances, we need to perform segmentation on multiple data sequences. If these sequences are not in relation, the problem is effortless since segmentation algorithms can be applied on each sequence independently. However, when the sequences are related to each other, performing multiple segmentation without considering relation among the sequences would induces inferior results. We take sequential segmentation and matching (SSM) problem as a study case. Given $m \geq 2$ data sequences, SSM aims at partitioning each sequence into several homogeneous segments and then establishing the correspondences between these segments from different sequences. A popular application of SSM is human action analysis. Specifically, the human action videos are segmented into primitive actions and the resulted sequences of the action segments are then aligned \cite{Qiu19,Chang19,Li20segali}.

\begin{figure*}[t]  
	\centering
	\subfigure[]{ \label{fig:CutoffMix_a}\includegraphics[width=0.95\textwidth]{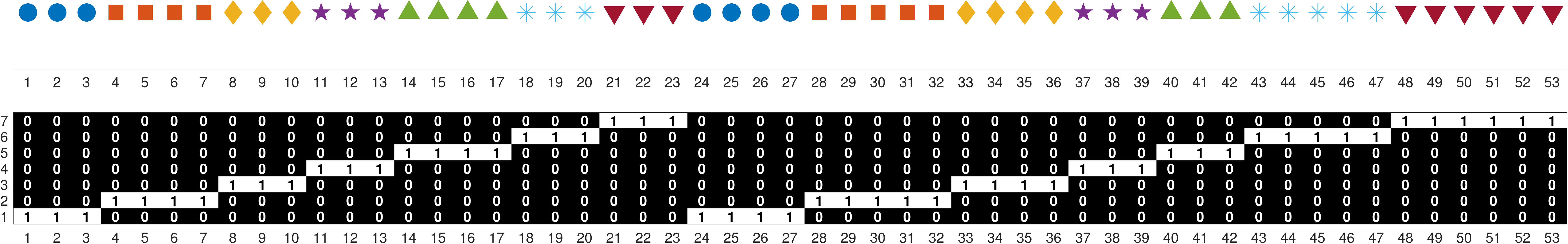} }
	\subfigure[]{ \label{fig:CutoffMix_b}\includegraphics[width=0.98\textwidth]{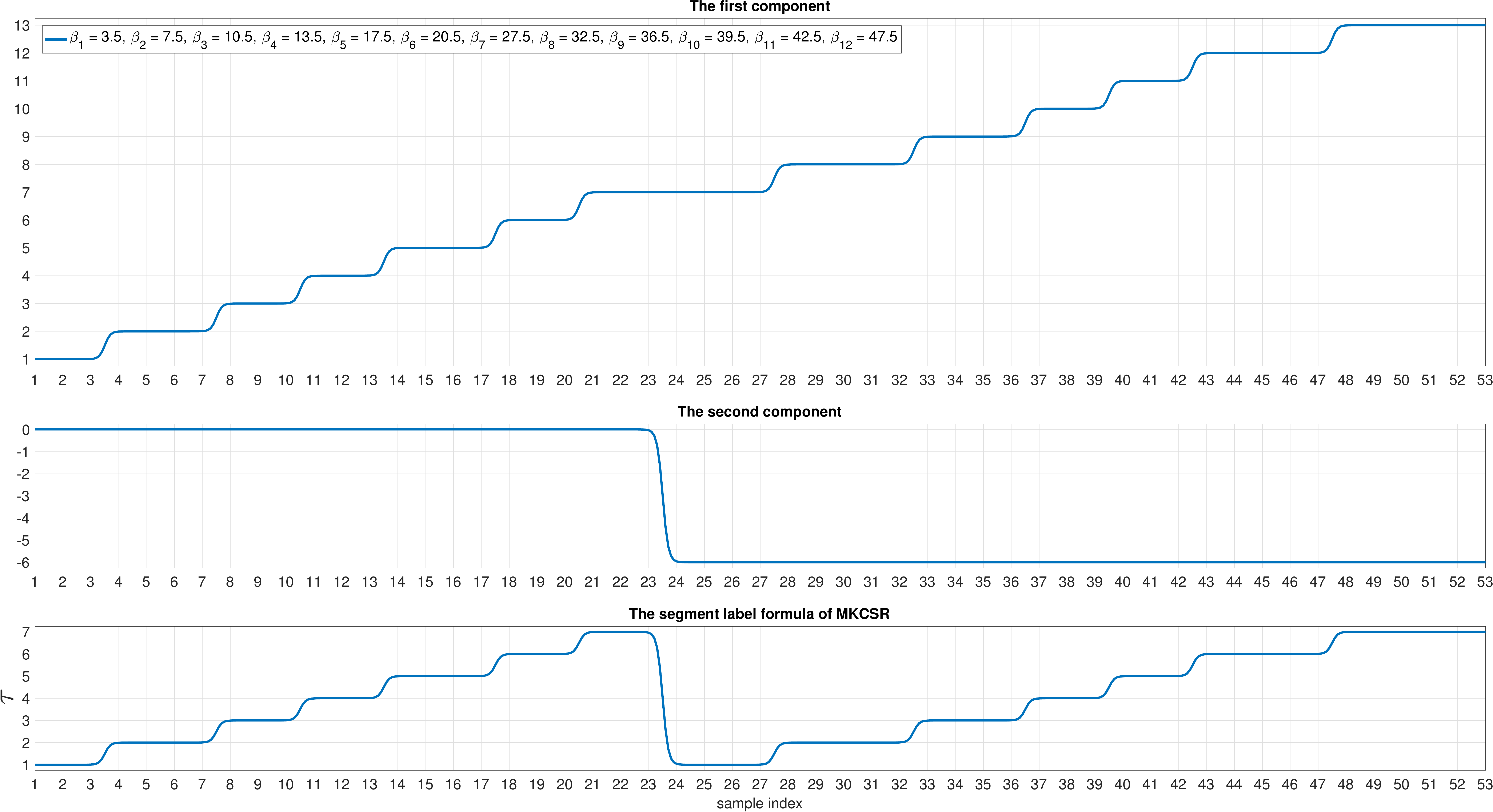} }
	\caption{ Illustration of the cut-off summation of sigmoid functions. (a) A toy example of a concatenation of two sequences ($m = 2, n_1 = 23, n_2 = 30$) and its corresponding indicator matrix ($k = 7$). (b) The cut-off summation of sigmoid functions, whose two components are depicts in the two first subfigures, can smoothly approximate the indicator matrix in the toy example.}
	\label{fig:CutoffMix}
\end{figure*}

To solve the SSM problem, in this work, we introduce an extension of the proposed model termed \textit{Multiple kernel clustering with sigmoid-based regularization} (MKSSR). MKSSR jointly partitions each data sequences into $k$ segments such that the $c^{\text{th}}$ segments of all the $m$ sequences are matched \footnote{Data samples of the $c^{\text{th}}$ segments from different sequences belong to the $c^{\text{th}}$ class for $1 \leq c \leq k$.}. Let $\bm{X}_p \in \mathbb{R}^{d \times n_p}$ for $1 \leq p \leq m$ denotes the $p^{\text{th}}$ data sequence and $\bm{G}_p \in \mathbb{R}^{k \times n_p}$ be its corresponding sample-to-segment indicator matrix. MKSSR firstly concatenates all the sequences to form a single long sequence $\bm{X} = [\bm{x}_1, \hdots, \bm{x}_m] \in \mathbb{R}^{d \times n}$, where $n = \sum_{i=1}^{m} n_p$. Then $\bm{G} = [\bm{G}_1, \hdots, \bm{G}_m] \in \mathbb{R}^{k \times n}$ is the corresponding indicator matrix of $\bm{X}$. Similar to the original KCSR, each element of $\bm{G}$ is defined as in (\ref{eq:G2tau}). However, in MKCSR, the segment label $\tau_j$ is computed as following
\begin{equation} \label{eq:mixture_cutoff}
	\begin{split}
		\tau_j = 1 & + \sum_{i=1}^{m(k-1)} f_{\text{sigmoid}}(j,\alpha,\beta_i) \\ & \qquad + (1-k) \sum_{p=1}^{m-1} f_{\text{sigmoid}}(j,\alpha,\sum_{q=1}^p n_q+0.5).
	\end{split}
\end{equation}
The function (\ref{eq:mixture_cutoff}), which we call as cut-off summation of sigmoid functions, consists of two components. The first component is the summation of sigmoid functions. It plays a similar role as (\ref{eq:mixture}) in KCSR. The second component presents the cut-off points (a.k.a junction points), at which two among the $m$ original data sequences are connected. It will reset the segment label from $k$ to $1$ after passing the final sample of one sequence and reaching a new sample from the next sequence. The cut-off summation of sigmoid functions and its components are illustrated in Figure \ref{fig:CutoffMix}.     

The formulation (\ref{eq:mixture_cutoff}) has $m(k-1)$ midpoint parameters, in which $\beta_{(p-1)(k-1) + 1}, \hdots, \beta_{p(k-1)}$ approximate the segment boundaries within the range $[1+\sum_{q=1}^{p-1} n_q,\sum_{q=1}^{p} n_q]$ for $1 \leq p \leq m$. Therefore, we introduce $mk$ parameters $\gamma_1, \hdots, \gamma_{mk}$ such that for $(p-1)(k-1) + 1 \leq j \leq p(k-1)$ 
\begin{equation} \label{eq:gamma_cutoff}
	\begin{split}
		\beta_i = \left( 1+\sum_{q=1}^{p-1} n_q \right) & \left(1 - \frac{\sum_{i\prime = (p-1)k + 1}^{i} e^{\gamma_{i^\prime}} }{\sum_{i^\prime = (p-1)k + 1}^{pk} e^{\gamma_{i^\prime}} }\right) \\
		& \qquad \; + \sum_{q=1}^{p} n_q \frac{\sum_{i\prime = (p-1)k + 1}^{i} e^{\gamma_{i^\prime}} }{\sum_{i^\prime = (p-1)k + 1}^{pk} e^{\gamma_{i^\prime}} }.
	\end{split}
\end{equation}
By replacing the last two constraints in (\ref{eq:obj_kcsr}) with (\ref{eq:mixture_cutoff}) and (\ref{eq:gamma_cutoff}) we can obtain the optimization problem of MKCSR. The objective function is then minimized w.r.t $mk$ parameters $\gamma_1, \hdots, \gamma_{mk}$ using the stochastic gradient descent algorithm.

\section{Experiments} \label{sec:experiments}

\subsection{Baselines}

We compare KCSR and its stochastic variant SKCSR with the following baselines

\begin{itemize}
	\item Aligned clustering analysis (ACA) \cite{Zhou13} -- a temporal clustering method that combines $k-$means with Dynamic time alignment kernel \cite{Shimodaira01}. 
	\item Sequential subspace clustering (SSC) \cite{Hu20} -- a temporal clustering method that combines subspace clustering with linearly temporal regularization weighted by $\ell_1-$norm sequential graph. 
	\item Temporal subspace clustering (TSC) \cite{Zheng21} -- a temporal clustering method that combines subspace clustering with manifold-based temporal regularization and low-rank constraint. 
	\item Approximate kernel segmentation (AKS) \cite{Celisse18} -- a heuristic approximation of the optimal kernel segmentation, where the solution is obtained by pruned DP algorithm that combines a low-rank approximation of the kernel matrix and the binary segmentation algorithm. 
	\item Greedy kernel segmentation (GKS) \cite{Truong19} -- another heuristic approximation of the optimal kernel segmentation that detects the segment boundaries sequentially using greedy algorithm.
\end{itemize}

\subsection{Datasets}

To evaluate performances of the above methods, we use a synthetic dataset and five real-world and widely public datasets.

\begin{figure*}[t]  
	\centering
	\subfigure[Data in $2D$]{ \label{fig:SynSeg_a}\includegraphics[width=0.3\textwidth]{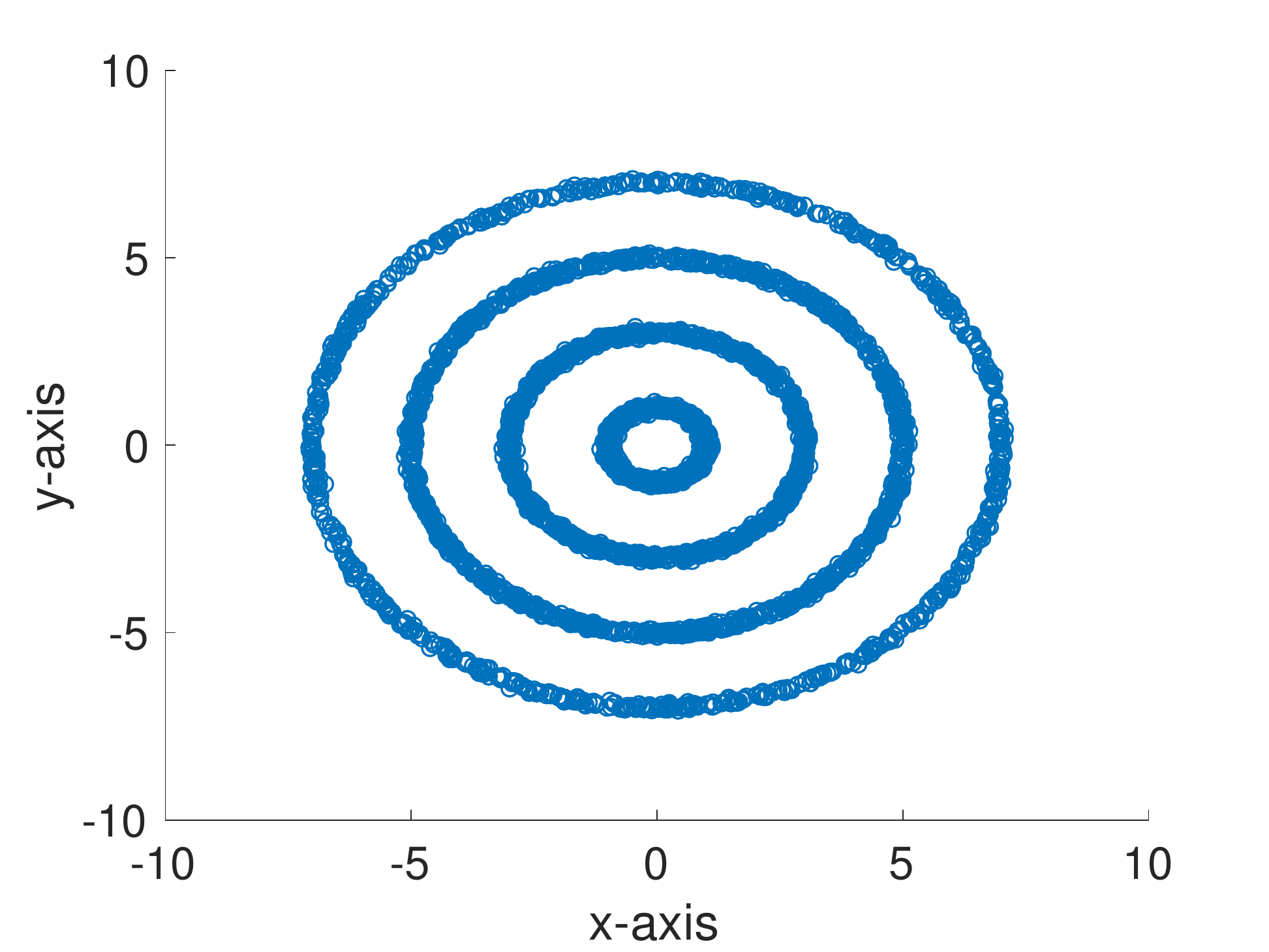} }
	\subfigure[Data in $3D$]{ \label{fig:SynSeg_b}\includegraphics[width=0.3\textwidth]{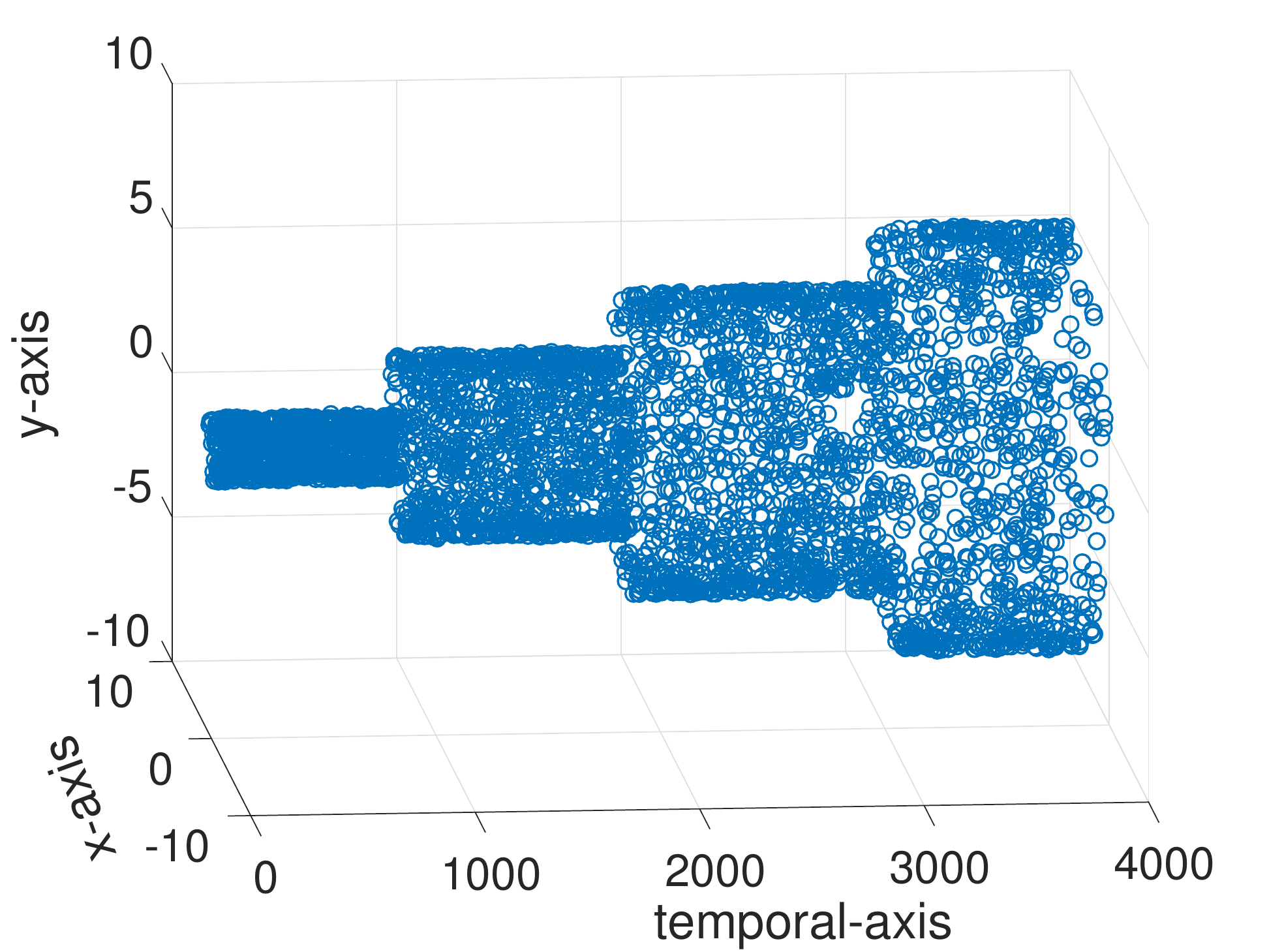} }
	\subfigure[SSC]{ \label{fig:SynSeg_c}\includegraphics[width=0.3\textwidth]{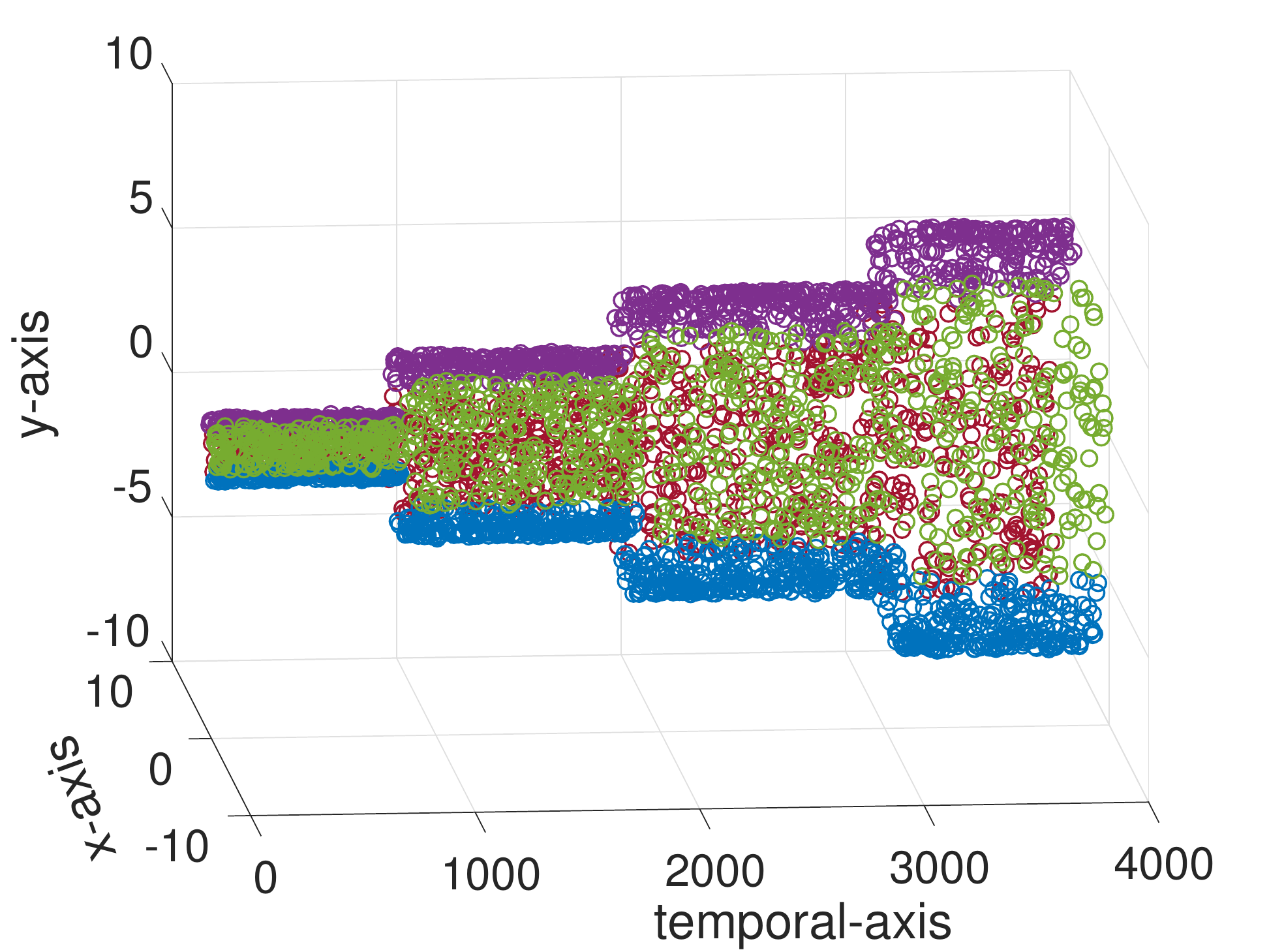} }
	\subfigure[TSC]{ \label{fig:SynSeg_d}\includegraphics[width=0.3\textwidth]{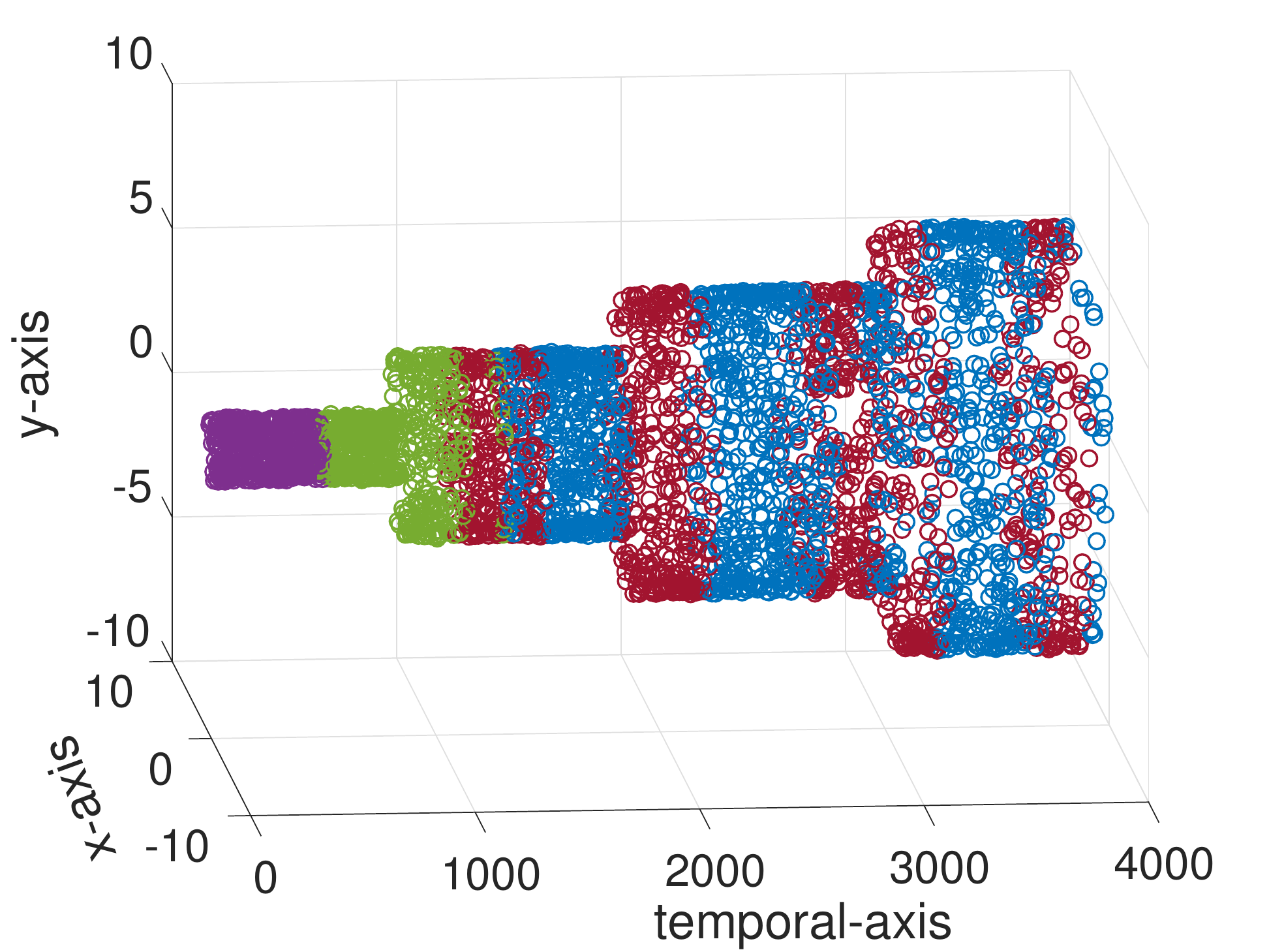} }
	\subfigure[ACA]{ \label{fig:SynSeg_e}\includegraphics[width=0.3\textwidth]{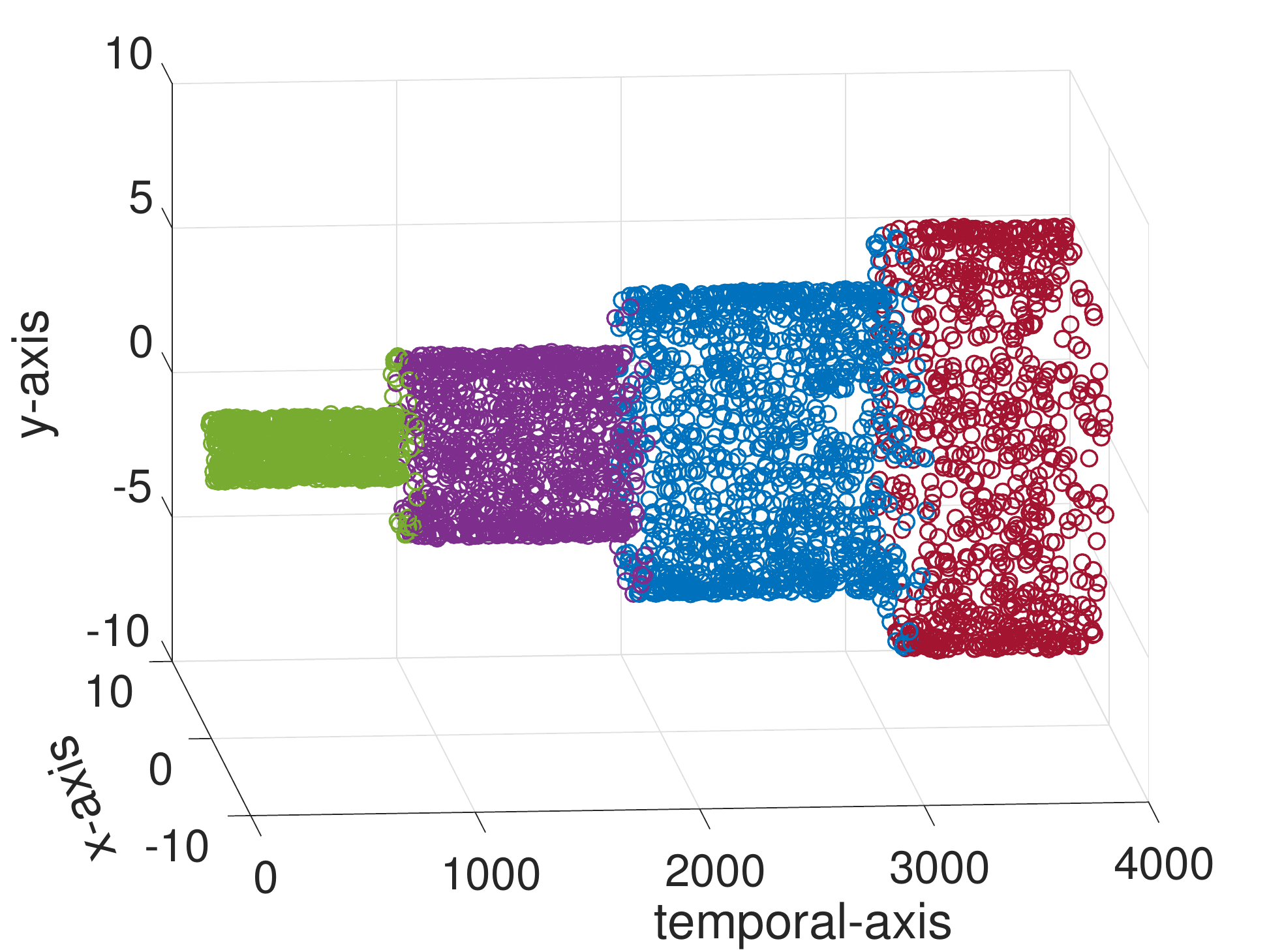} }
	\subfigure[AKS]{ \label{fig:SynSeg_f}\includegraphics[width=0.3\textwidth]{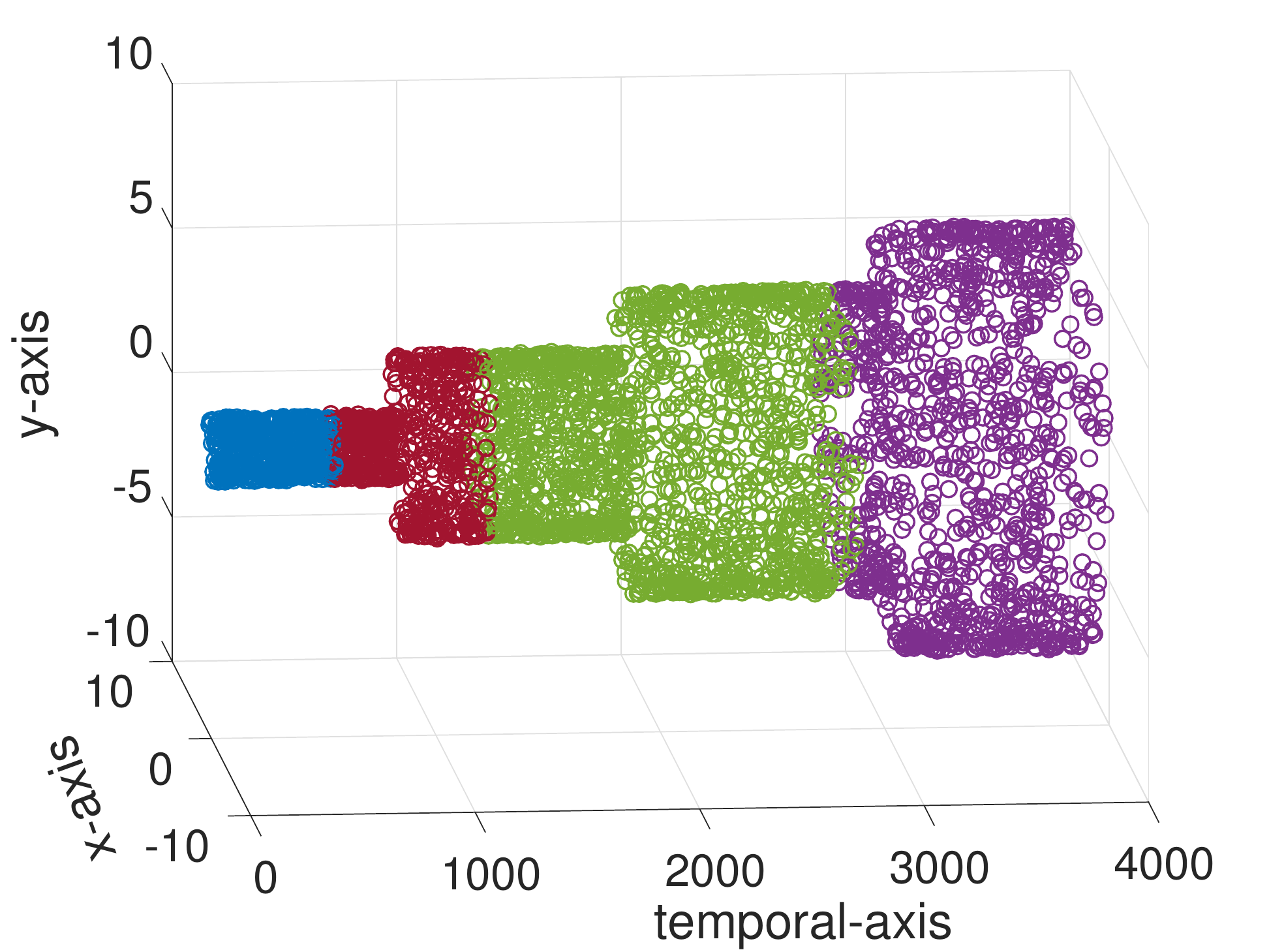} }
	\subfigure[GKS]{ \label{fig:SynSeg_g}\includegraphics[width=0.3\textwidth]{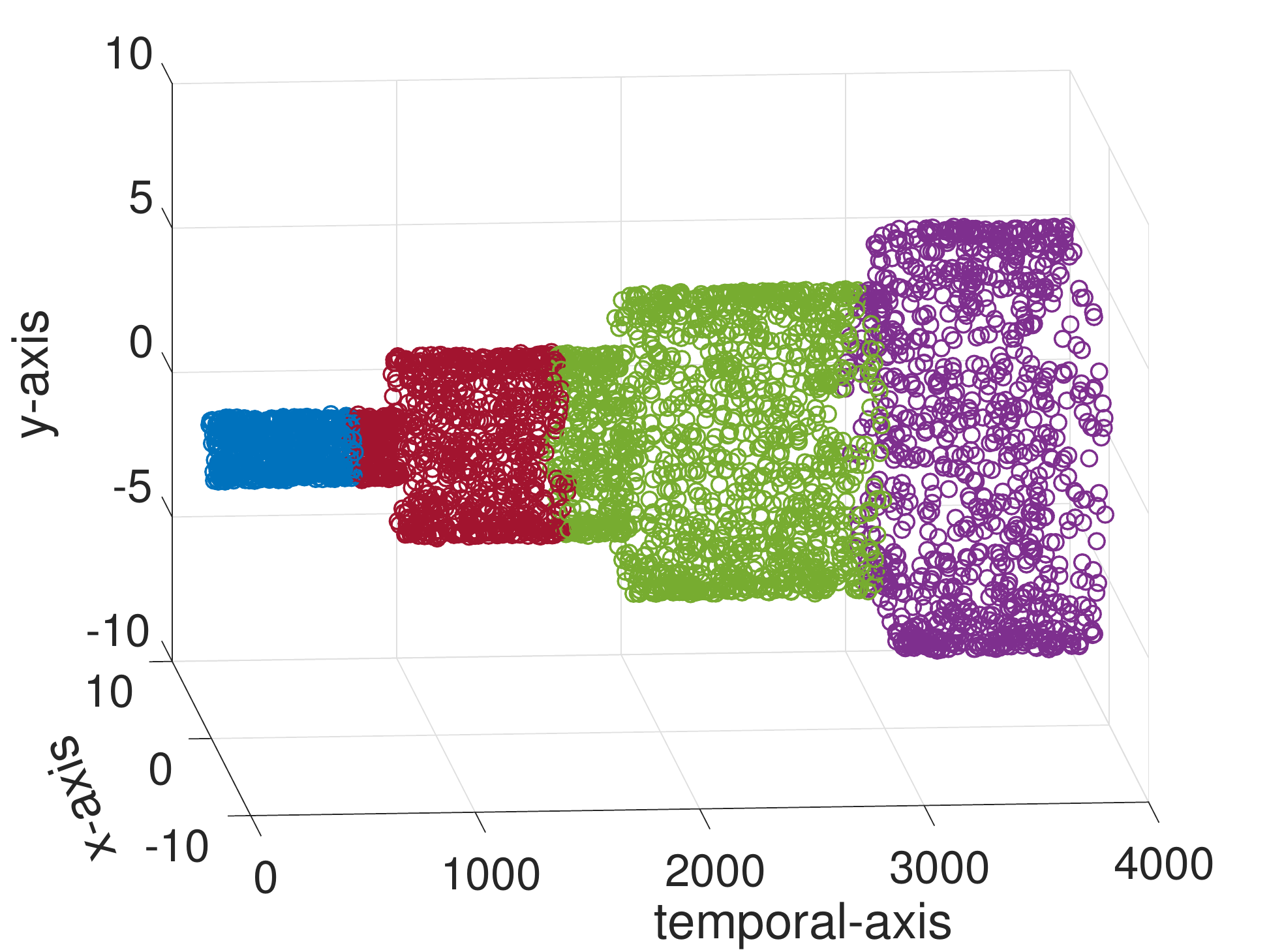} }
	\subfigure[KCSR]{ \label{fig:SynSeg_h}\includegraphics[width=0.3\textwidth]{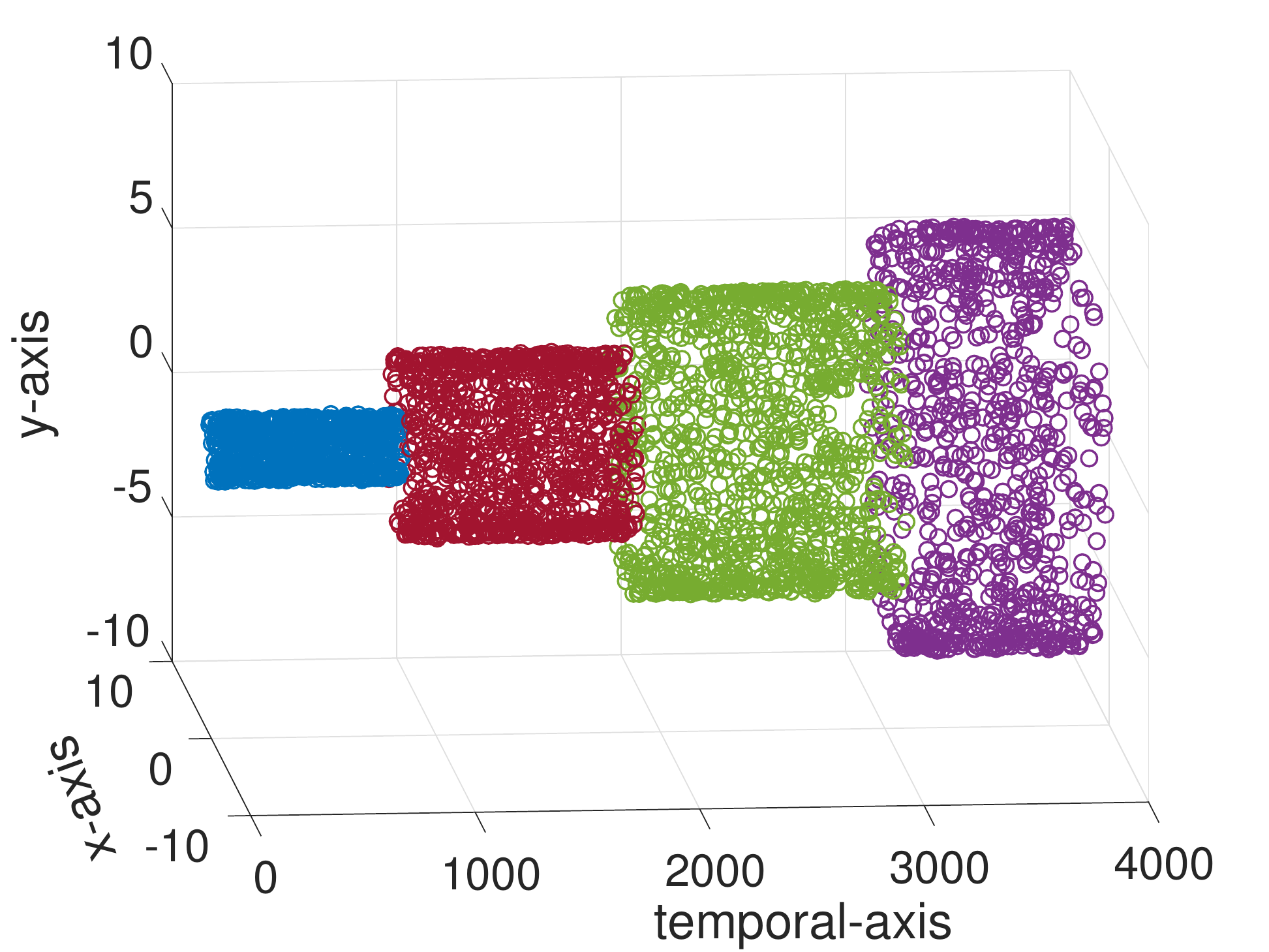} }
	\subfigure[SKCSR]{ \label{fig:SynSeg_i}\includegraphics[width=0.3\textwidth]{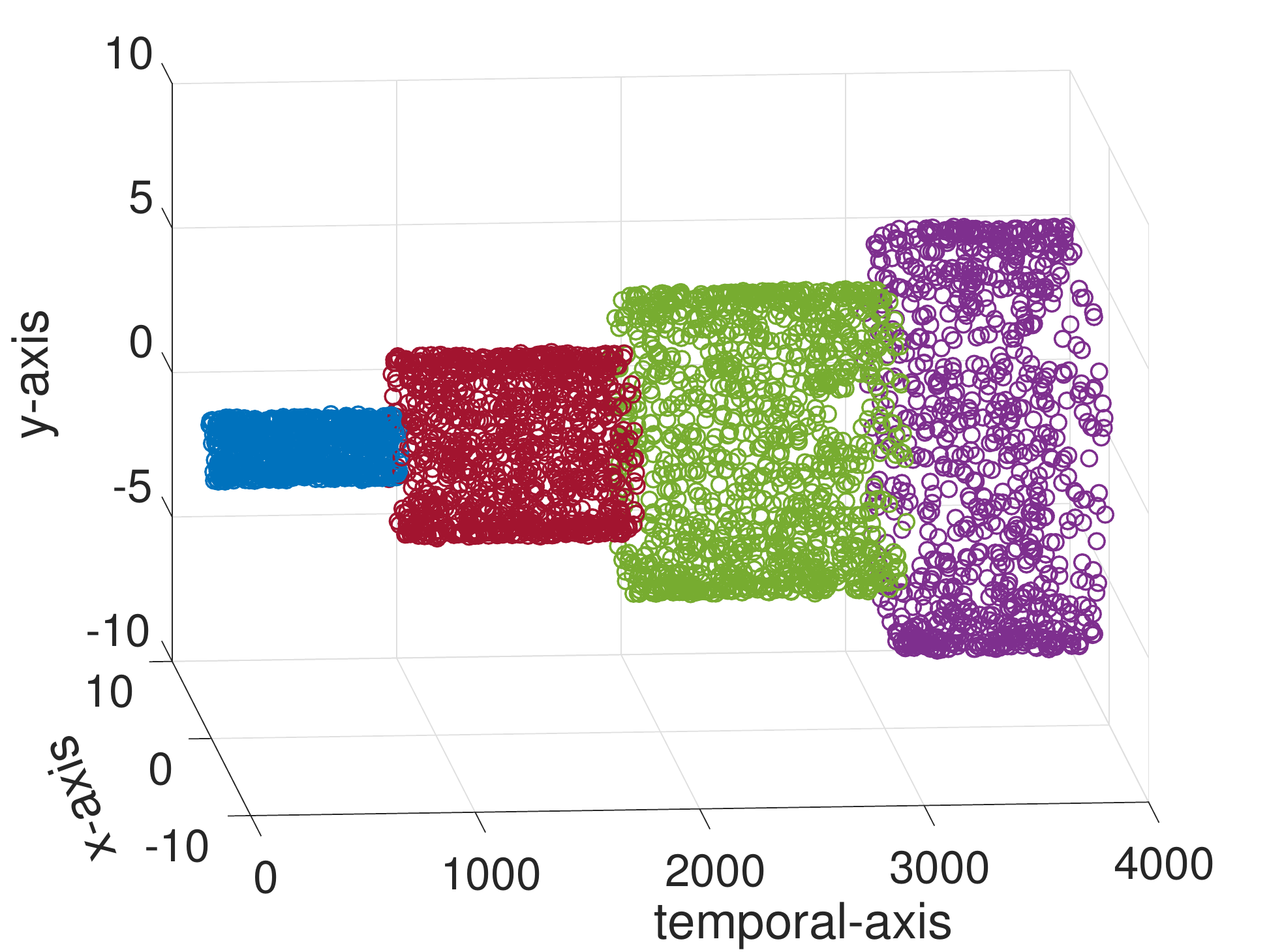} }
	\caption{ Synthetic experiment: (a) data generated in $2D$ space, (b) the data after contiguously rearranging and visualization of segmentation results returned by all the compared methods. Different colors represent different clusters.}
	\label{fig:SynSeg}
\end{figure*}

\textbf{Synthetic data.} We first generate $2D$ data samples that form four circles of different diameters. They are illustrated in Figure \ref{fig:SynSeg_a}. The number of data samples of each circle is randomly selected in range $[500,1500]$ and also constrained to be different. For instance, in our case, the numbers of data samples of the circles from low to high diameters are $832, 1018, 1174$ and $843$, respectively. We then rearrange the generated data samples in contiguous order, i.e. data samples of one circle do not mix to the other circles. By doing so, each circle in the original $2D$ space corresponds to a segment in the new sequential data. See Figure \ref{fig:SynSeg_b} for illustration.

\begin{figure}[t]
	\centering
	\subfigure[Weizmann data]{ \label{fig:WeiMniData_a}\includegraphics[width=0.46\textwidth]{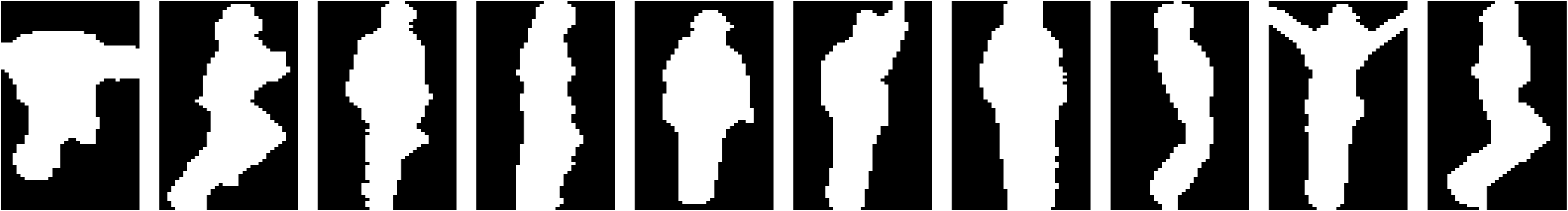} }
	\subfigure[Ordered MNIST data]{ \label{fig:WeiMniData_b}\includegraphics[width=0.46\textwidth]{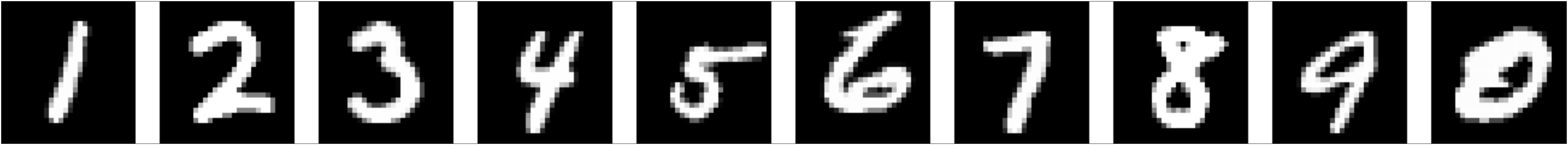} }
	\caption{(a) Concatenated action videos of subject $1$ in Weizmann dataset and (b) the rearranged digit images sequence in MNIST dataset. Each data sequence consists of $10$ non-overlapping segments and only one representative frame of each segment is depicted.}
\end{figure}

\textbf{Weizmann data.} The Weizmann dataset \cite{Gorelick07} consists of $90$ videos of nine subjects, each performing ten actions: bend, run, jump-in-place (pjump), walk, jack, wave-one-hand (wave1), side, jump-forward (jump), wave-two-hand (wave2), and skip. Similar to \cite{Hoai14}, videos of the same subjects are concatenated into a long video sequence following the presented order of the actions. We then subtract background from each frame of these video sequences and rescale them to the size $70 \times 35$. For each $70-$by$-35$ rescaled frame, we compute the binary feature as shown in Figure \ref{fig:WeiMniData_a}. To reduce the dimensions of the feature space (2450), the top 123 principal components that preserve $99\%$ of the total energy are kept for experiments.

\begin{figure*}[t]
	\centering
	\includegraphics[width=0.8\textwidth]{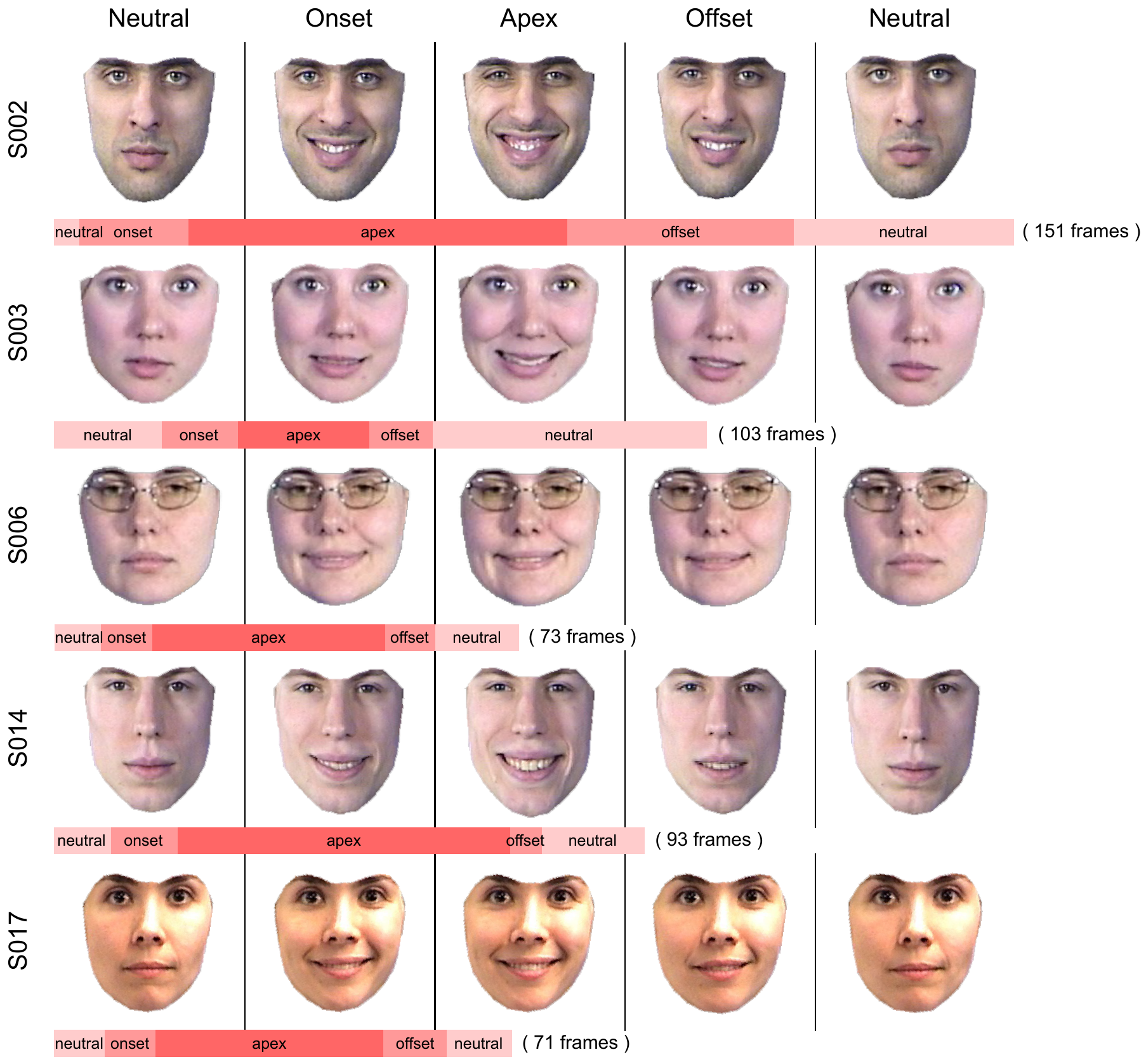}
	\caption{Videos of five subjects (S002, S003, S006, S014 and S017) performing an action unit $12$ that corresponds to smile taken from MMI Facial action units dataset. The representative facial images of the segments are depicted. The bottom of each video shows duration of the corresponding ground truth frame-to-segment labels along with the total number of frames.}
	\label{fig:MMI}
\end{figure*}

\textbf{MMI Facial action units.} We exploit the MMI Facial Expression dataset \cite{Pantic05}, which contains more than $2900$ videos of $75$ different subjects, each performing a particular combination of Action Unit (AU). In this paper, we focus on videos of AU12, which corresponds to a smile. Although, these videos consist of different number of frames, they are composed of exactly five segments with the following order: neutral, onset, apex, offset, neutral, where \textit{neutral} is when facial muscle is inactive, \textit{apex} is when facial muscle intensity is strongest, and \textit{onset} is when facial muscle starts to activate or \textit{offset} is when facial muscle begins to relax. Following the same pre-processing procedure as in \cite{Doan20}, we cropped and aligned the face using dlib-ml \cite{King09}. The results are depicted in Figure \ref{fig:MMI}. We then convert them to grayscale and reduce their dimension to $400$ using whitening PCA. We finally selected videos of five subjects $2,3,6,14$ and $17$ for experiments. Their ground-truth frames-to-segment lables are already given in the original dataset.

\textbf{Google spoken digits.}  Google's Speech Commands (GSC) \cite{Warden18, McMahan18} is a large audio dataset that consists of more than $30$ categories of spoken terms. For each category that relates to digits from ``one'' to ``nine'', we randomly select a clean recording. These recordings are then concatenated, forming a long audio sequence with $19$ segments ($9$ segments of active voice and $10$ silent segments) (see Fig. \ref{fig:GooSeg}). We further add white noise, which is also provided in the GSC dataset, to make the segmentation problem more challenging. Finally, a sequence of acoustic features, which are 13-dimensional mel--frequency cepstral coefficients (MFCCs) \cite{Davis80} for every $10$ms of a $25$ms window, is computed from the noisy audio sequence. The annotation is manually obtained based on the log filter-bank energies of the clean audio. 

\textbf{Ordered MNIST data.} the MNIST dataset \cite{LeCun98} consists of $28 \times 28$ grayscale digit [0, 9] images divided into $60K/10K$ for training/testing. Since all the compared methods are unsupervised and require no training phase, we use all $70K$ images to perform segmentation. Note that the original data is not exact suited to the sequential assumption. Following the same setting of \cite{Zheng21}, we rearrange order of the images such that those of the same digit form a contiguous segment and the ten segments are concatenated into a very long images sequence (see Figure \ref{fig:WeiMniData_b}). Different from \cite{Zheng21}, where only $2K$ images were selected for experiment, our ordered MNIST data consists of the whole $70K$ images. To handle this large-scale data, temporal clustering and kernel CPD methods requires up to $36.5$ GB to store the kernel and/or affinity graph matrices, which is impractical for implementation on a single personal PC. Among the compared methods, only SKCSR and AKS with low memory complexities can perform segmentation on this dataset. 

\textbf{Acceleration data.}\footnote{The acceleration dataset is available in UCI repositories and can be retrieved from \url{https://archive.ics.uci.edu/ml/machine-learning-databases/00287/} .} The acceleration data \cite{Casale12} are acquired from a triaxial accelerometer mounted on the chests of $15$ subjects, each performing a sequence of activities such as working at computer, standing, walking, going updown stairs and talking. The aims of our experiments is to partition the data sequences into segments that correspond to the activities. Thus, we firstly pre-process the data. For each subject, we add squares of signals from the three axles. An example is depicted in Figure \ref{fig:AccSeg}. We then transform the obtained summation signal using wavelet transform with scale factor $64$ and the Morlet wavelet as the mother wavelet function. The resulting 2D wavelet coefficient matrix $\bm{C}$ is of the size $64$-by-$n_{acc}$, where $n_{acc}$ is the length of the original acceleration signal. Note that the wavelet coefficients $C$ are complex numbers. Thus we take its modulus as input for the methods in our experiments. Similar to the Ordered MNIST data, this dataset consist of long acceleration sequences. The average $n_{acc}$ is $125K$. Therefore, the methods with memory complexities of order $O(n^2)$ will require up to approximately $116.4$ GB, which is unaffordable in our case, for storage. In our experiments, only SKCSR and AKS can handle this dataset.

\begin{table*}[t]
	\begin{center}
		\resizebox{\textwidth}{!}{
			\begin{tabular}{|c|c|c|c|c|c|c|c|c|}
				\hline
				\multicolumn{2}{|c|}{Dataset} & SSC & TSC & ACA & AKS & GKS & KCSR & SKCSR \\
				\hline \hline
				\multirow{2}{*}{\textbf{Synthetic data}} & \mycc ACC & \mycc $0.0965 \; (0.0736)$ & \mycc $0.4077 \; (0.0795)$  & \mycc $0.9305 \; (0.0224)$ & \mycc $0.6577 \; (0.0671)$ & \mycc $0.6999 \; (0.0255)$ & \mycc $\bm{0.9871} \; (0.0104)$ & \mycc $0.9870 \; (0.0092)$ \\
				\cline{2-9}
				& NMI & $0.1070 \; (0.0649)$ & $0.3608 \; (0.0758)$ & $0.9214 \; (0.0375)$ & $0.6067 \; (0.0751)$ & $0.7036 \; (0.0325)$ & $\bm{0.9959} \; (0.0024)$ & $0.9847 \; (0.0077)$ \\
				\hline
				\multirow{2}{*}{\textbf{Weizmann}} & \mycc ACC & \mycc $0.4856 \; (0.0269)$ & \mycc $0.7028 \; (0.0430)$  & \mycc $0.7687 \; (0.0146)$  & \mycc $0.7182 \; (0.0188)$ & \mycc $0.5628 \; (0.0218)$ & \mycc $0.8835 \; (0.0092)$ & \mycc $\bm{0.8964} \; (0.0113)$ \\
				\cline{2-9}
				& NMI & $0.4157 \; (0.0387)$ & $0.7207 \; (0.0501)$ & $0.7628 \; (0.0196)$ & $0.7006 \; (0.0205)$ & $0.6032 \; (0.0262)$ & $0.9071 \; (0.0107)$ & $\bm{0.9151} \; (0.0095)$ \\
				\hline
				\multirow{2}{*}{\textbf{MMI Facial AU}} & \mycc ACC & \mycc $0.6453 \; (0.0238)$ & \mycc $0.7552 \; (0.0392)$ &  \mycc $0.8121 \; (0.0157)$ & \mycc $0.7527 \; (0.0204)$ & \mycc $0.6025 \; (0.0249)$ & \mycc $0.9537 \; (0.0118)$ & \mycc $\bm{0.9763} \; (0.0075)$ \\
				\cline{2-9}
				& NMI & $0.6514 \; (0.0412)$ & $0.7321 \; (0.0437)$ & $0.7952 \; (0.0205)$ & $0.7600 \; (0.0214)$ & $0.6355 \; (0.0298)$ & $0.9625 \; (0.0129)$ & $\bm{0.9686} \; (0.0126)$ \\
				\hline
				\multirow{2}{*}{\textbf{Google}} & \mycc ACC & \mycc $0.2057 \; (0.0275)$ & \mycc $0.6434 \; (0.0313)$ & \mycc $0.8241 \; (0.0182)$ & \mycc $0.6726 \; (0.0257)$ & \mycc $0.7458 \; (0.0294)$ & \mycc $0.7914 \; (0.0172)$ & \mycc $\bm{0.8826} \; (0.0165)$ \\
				\cline{2-9}
				& NMI & $0.1839 \; (0.0256)$ & $0.6513 \; (0.0330)$ & $0.7939 \; (0.0196)$ & $0.6954 \; (0.0266)$ & $0.7557 \; (0.0305)$ & $0.8109 \; (0.0154)$ & $\bm{0.9009} \; (0.0186)$ \\
				\hline
				\multirow{2}{*}{\textbf{Ordered MNIST}} & \mycc ACC & \mycc -- & \mycc -- & \mycc -- & \mycc $0.6983 \; (0.0282)$ & \mycc -- & \mycc -- & \mycc $\bm{0.9681} \; (0.0155)$ \\
				\cline{2-9}
				& NMI & -- & -- & -- & $0.7196 \; (0.0209)$ & -- & -- & $\bm{0.9819} \; (0.0119)$ \\
				\hline
				\multirow{2}{*}{\textbf{Acceleration}} & \mycc ACC & \mycc -- & \mycc -- & \mycc -- & \mycc $0.5535 \; (0.0434)$ & \mycc -- & \mycc -- & \mycc $\bm{0.8172} \; (0.0216)$ \\
				\cline{2-9}
				& NMI & -- & -- & -- & $0.5763 \; (0.0398)$ & -- & -- & $\bm{0.8056} \; (0.0311)$ \\
				\hline
			\end{tabular}
		}
	\end{center}
	\caption{Segmentation results on six datasets, including synthetic data, Weizmann action sequences, MMI Facial smiling video, noisy Google spoken digits, ordered MNIST data and Acceleration sequences, returned by different methods. The mean score of each methods over five random runs along with its variance are reported. The symbol ``--'' means that there is no result due to the shortage of memory resources. }
	\label{tab:result_scores}
\end{table*}

 \subsection{Evaluation measures}
 
 Given a specific value $k$, while KCSR, SKCSR, AKS and GKS return exactly $k$ non-overlapping segments, temporal clustering-based methods partition samples of the data sequence into $k$ clusters that maybe dispersed in discontiguous segments. Since, all the compared methods base on clustering scheme, we use accuracy and normalized mutual information \cite{Cai05} as evaluation metrics to assess the segmentation results.
 
 Let $\widehat{\bm{\mathcal{L}}} = [\hat{l}_1, \hdots, \hat{l}_n]$ and $\bm{\mathcal{L}} = [l_1, \hdots, l_n]$ be the obtained labels and ground-truth labels of a given data sequence $\bm{X} = \left[ \bm{x}_1, \hdots, \bm{x}_n \right]$. $\hat{l}_j = i$ (similar for $l_j$) for $1 \leq i \leq k$ indicates that $\bm{x}_j$ belongs to cluster (segment) $\hat{c}_i$. The accuracy (ACC) is defined as follows:
 \begin{equation}
 	ACC = \frac{\sum_{j=1}^{n} \delta(l_j,map(\hat{l}_j))}{n},
 \end{equation}
 where $\delta(a,b)$ is the delta function that equals one if $a = b$ and zero otherwise and $map(\hat{l}_j)$ is the permutation mapping  function that maps label $\hat{l}_j$ to the equivalent ground truth label. In this work, we use Kuhn-Munkres algorithm \cite{Lovasz86} to find the mapping. 
 
 Let $\widehat{\mathcal{C}} = [\hat{c}_1, \hdots, \hat{c}_k]$ and $\mathcal{C} = [c_1, \hdots, c_k]$ be the obtained clusters and the ground-truth clusters. Their mutual information (MI) is
 \begin{equation}
 	MI(\mathcal{C},\widehat{\mathcal{C}}) = \sum_{c_i \in \mathcal{C}, \hat{c}_{i^\prime} \in \widehat{\mathcal{C}}} p(c_i,\hat{c}_{i^\prime}) \log_2 \frac{p(c_i,\hat{c}_{i^\prime})}{p(c_i) p(\hat{c}_{i^\prime})},
 \end{equation}
 where $p(c_i)$ and $p(\hat{c}_{i^\prime})$ are the probabilities that a data sample arbitrarily selected from the sequence belongs to the clusters $c_i$ and $\hat{c}_{i^\prime}$, respectively, and $p(c_i,\hat{c}_{i^\prime})$ is the joint probability that the selected data sample belongs to both $c_i$ and $\hat{c}_{i^\prime}$. This metric is normalized to the range $[0,1]$ as follows:
 \begin{equation}
 	NMI(\mathcal{C},\widehat{\mathcal{C}}) = \frac{MI(\mathcal{C},\widehat{\mathcal{C}})}{max ( H(\mathcal{C}), H(\widehat{\mathcal{C}}) )},
 \end{equation}
 where $H(\mathcal{C})$ and $H(\widehat{\mathcal{C}})$ are the entropies of $\mathcal{C}$ and $\widehat{\mathcal{C}}$, respectively.
 
 \subsection{Parameter settings} \label{sec:exp_params}
 
 We select the optimal parameters for each method to achieve the best performance. The number of clusters $k$ of all the compared methods is set to the number of segments available in the datasets. For ACA, its parameters $nMa$ and $nMi$ that specify the maximum and minimum lengths of each divided subsequence, respectively, are data-dependent. Let $n$ be the sequence length, we select $nMa$ from a rounded set $\{0.01n, 0.02n, 0.04n, 0.06n, 0.08n, 0.1n\}$ and set $nMi = \frac{nMa}{2}$. For temporal subspace clustering methods, including SSC and TSC, the most important parameter is that controls the sequential regularization for the new representation $\bm{Z}$. We select this parameter from the set $\{1,5,10,15,20,25\}$ and the other parameters are set according to the original papers. For the proposed methods, we fix the parameter that controls the steepness of the summation of sigmoid functions at the midpoints $\alpha = 10$. The tolerance $\epsilon$ for convergence verification in KCSR is fixed at $10^{-6}$. For all the datasets, we use the Radial Basis Function (RBF) Kernel \footnote{ RBF kernel: $\kappa(\bm{x}_i,\bm{x}_j) = \e^{-\frac{\lvert \lvert \bm{x}_i - \bm{x}_j \rvert \rvert^2_2}{2 \sigma^2}}$.} with proper width $\sigma$ for AKS, GKS and the proposed methods. The minibatch size $b$ of SKCSR and the rank $r$ of the approximation of the kernel matrix in AKS are kept equal. Their values are selected from a set $\{64, 128, 256, 512, 1024, 2048\}$. Note that, SKCSR terminates after processing $T$ minibatches. We set $T$ such that $T \times b \geq 50n$ (passing through the data sequence at least $50$ times). \textit{}

\begin{figure}[t] 
	\centering
	\includegraphics[width=0.46\textwidth]{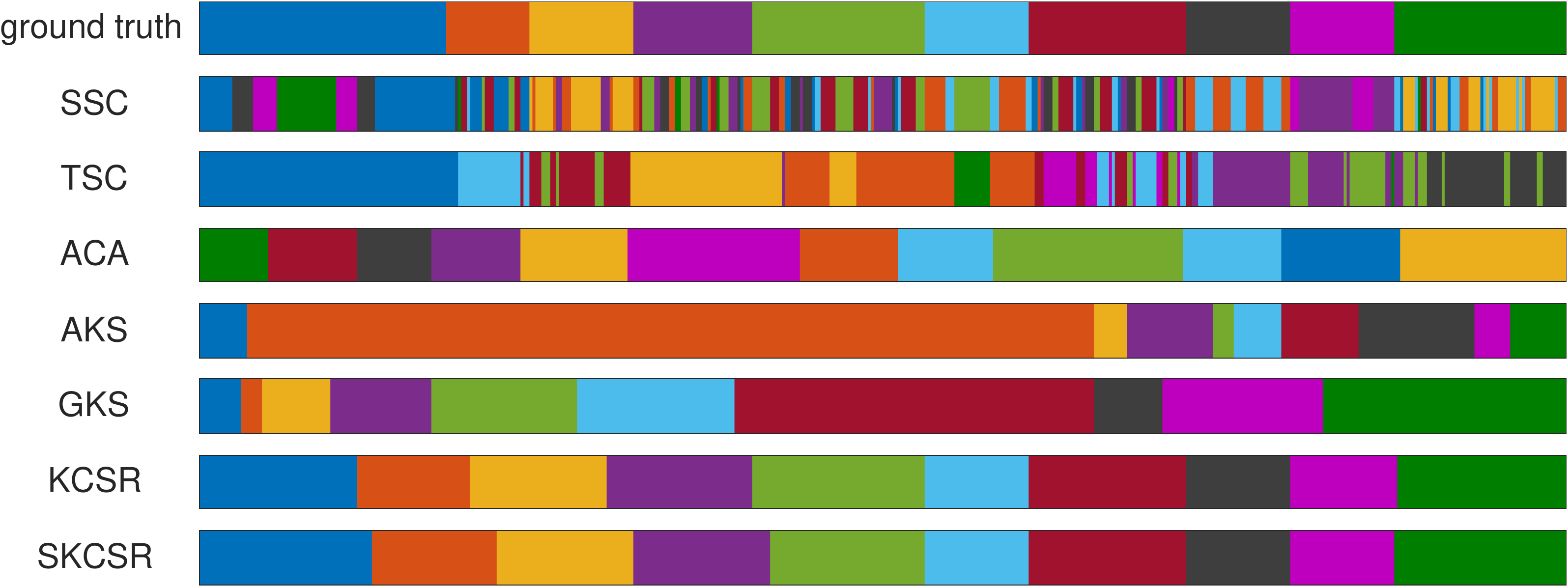}
	\caption{Visualization of segmentation results returned by the proposed methods and baselines on Weizmann dataset. Different colors represent different clusters.}
	\label{fig:WeiSeg}
\end{figure}

\begin{figure}[t]
	\centering
	\includegraphics[width=0.46\textwidth]{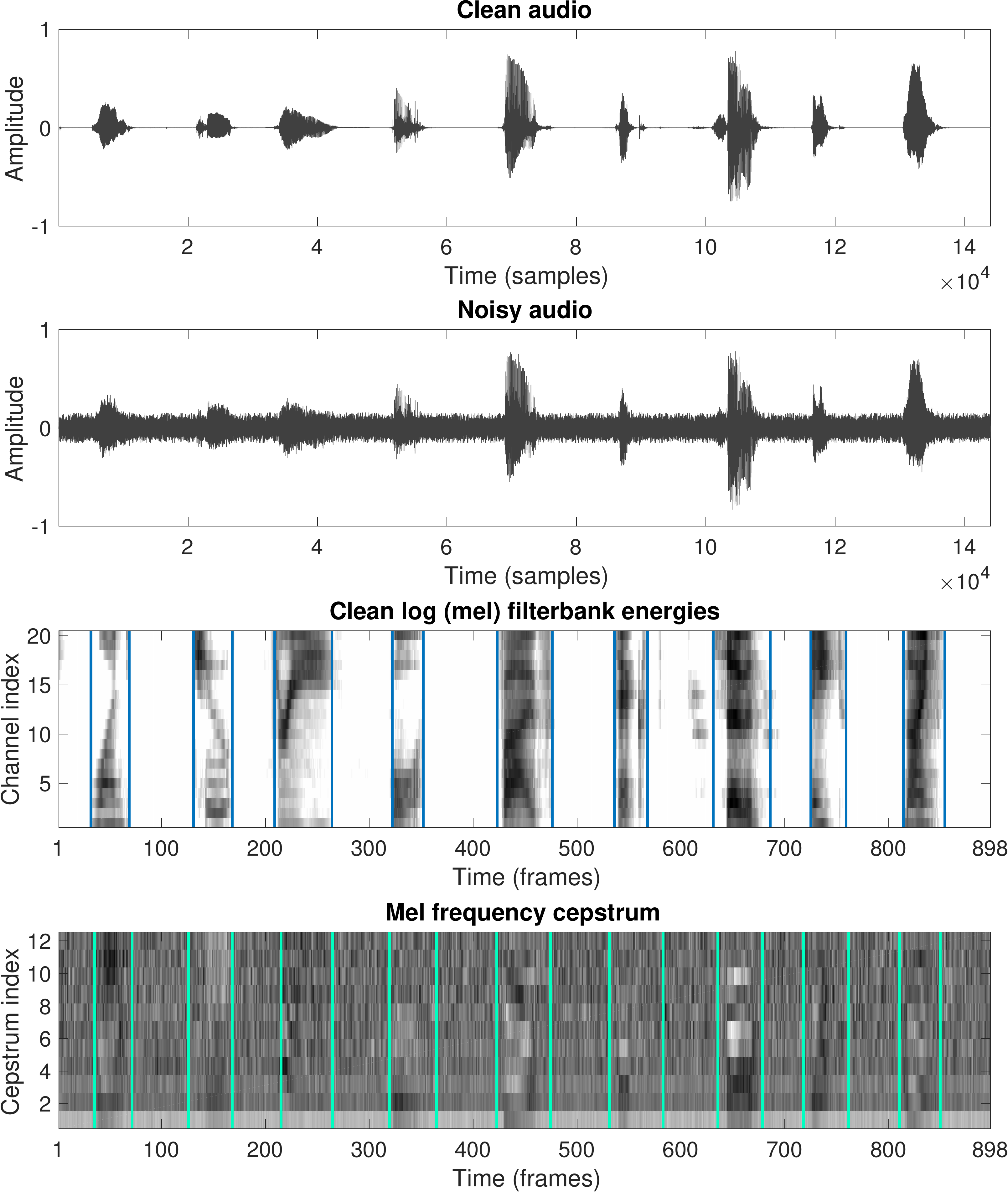}
	\caption{From the top to the bottom: clean audio of spoken digits [1,9], the audio contaminated by white noise, log filter-bank energies of the clean audio used for manual annotation (blue lines depict ground truth segment boundaries) and Mel-frequency cepstrum of the noisy audio (vertical lines show the midpoints $\bm{\beta}$ of the summation of sigmoid functions returned by SKCSR).}
	\label{fig:GooSeg}
\end{figure}

\begin{figure}[t] 
	\centering
	\includegraphics[width=0.46\textwidth]{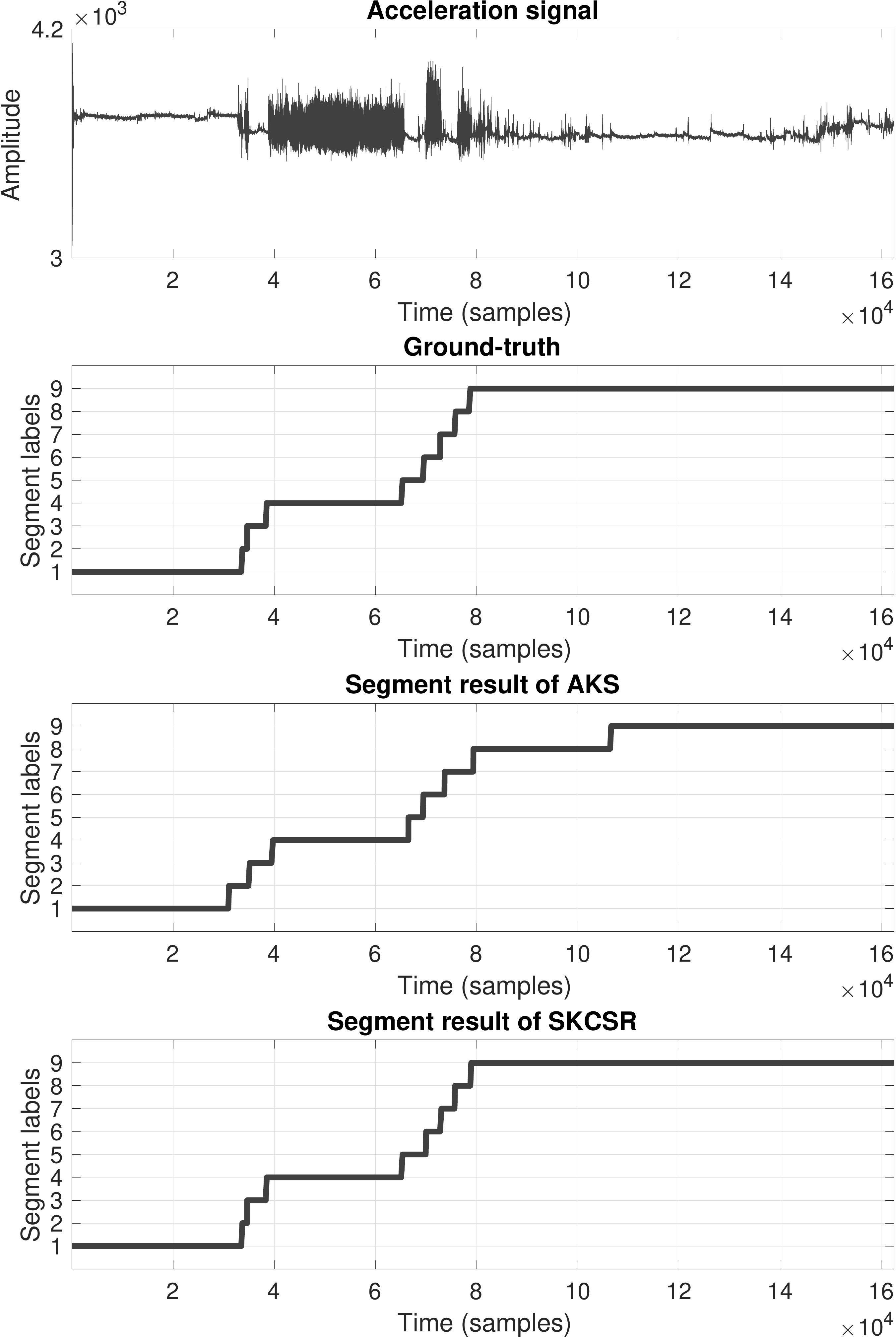}
	\caption{From the top to the bottom: Acceleration signal of the first subject, the corresponding ground-truth segment labels and the segmentation results returned by AKS and SKCSR, respectively, on the Acceleration dataset. }
	\label{fig:AccSeg}
\end{figure}

\subsection{Results discussion}

\subsubsection{Evaluation of KCSR and SKCSR}

Figure \ref{fig:SynSeg} visualizes the segmentation results on synthetic data and the evaluation scores are given in the first rows of Table \ref{tab:result_scores}. We can observe that each segment of the generated data sequence has a circular structure. Therefore, the nonlinear regularization in sequential representation learning of SSC is ineffective on the synthetic data. TSC performed significantly better. The manifold-based regularization allows it to be able to capture the nonlinear structure in the data. Our methods also perform segmentation based on regularization. However, different from SSC and TSC, where the regularization is just local\footnote{The regularization only preserves the local relationship on representation of consecutive samples.}, the summation of sigmoid functions of KCSR and SKCSR globally regularizes the whole data sequences and the locality is ensured by its smooth nature. Therefore, the proposed methods obtained the best performance on the synthetic dataset.

On the real-world data, including Weizmann action videos, MMI Facial smiling videos and Google spoken digits audio, the proposed models also outperformed the baselines.  Evaluation scores of the corresponding segmentation results are shown in the second, third and fourth rows of Table \ref{tab:result_scores}. We can observe that ACA also had good performances on these datasets. Although ACA also performs segmentation based on clustering as our methods do, it cannot guarantee to find exact $k$ non-overlapping segments. Therefore, its evaluation scores are slightly lower than those of the proposed models. In comparison with heuristic approximations AKS and GKS, our models also had better performances. Similar to AKS and GKS, our models also search for segment boundaries. They approximate the boundaries by midpoints $\bm{\beta}$ of the summation of sigmoid functions. However, different from these heuristic approximations that search for the segment boundaries sequentially, the proposed models simultaneously obtain all the $\bm{\beta}$ via gradient-based algorithm. As convergence of this optimization algorithm is theoretically proved, optimality of the solutions is guaranteed. To qualitatively assess the performances of the compared methods, we also visualized the segmentation results on Weizmann video and Google audio datasets in Figure \ref{fig:WeiSeg} and Figure \ref{fig:GooSeg}, respectively. These visualization further validate the superior performances of our methods over those of the baselines. 

\begin{figure*}[t]
	\centering
	\includegraphics[width=0.98\textwidth]{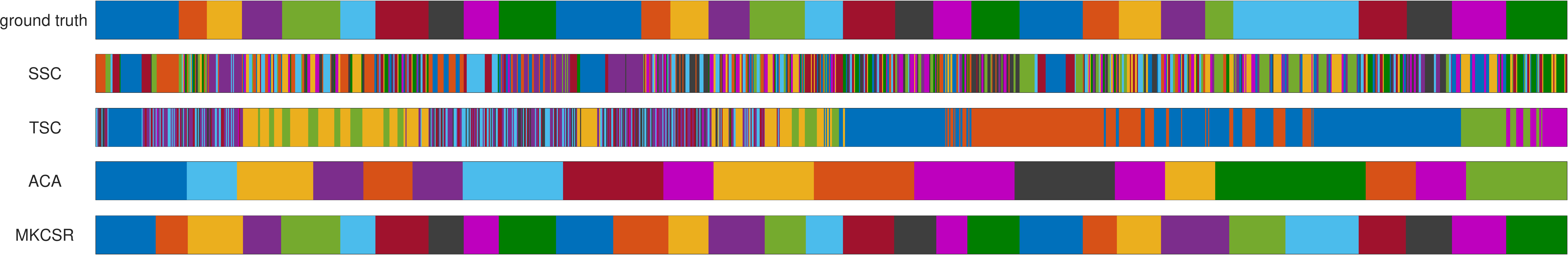}
	\caption{Visualization of segmentation results of SSC, TSC, ACA and MKCSR on three concatenated action video sequences from Weizmann dataset.}
	\label{fig:MulSeg}
\end{figure*}

\begin{figure}[t] 
	\centering
	\subfigure[Weizmann]{\label{fig:WeiMniCon_a}\includegraphics[width=0.46\textwidth]{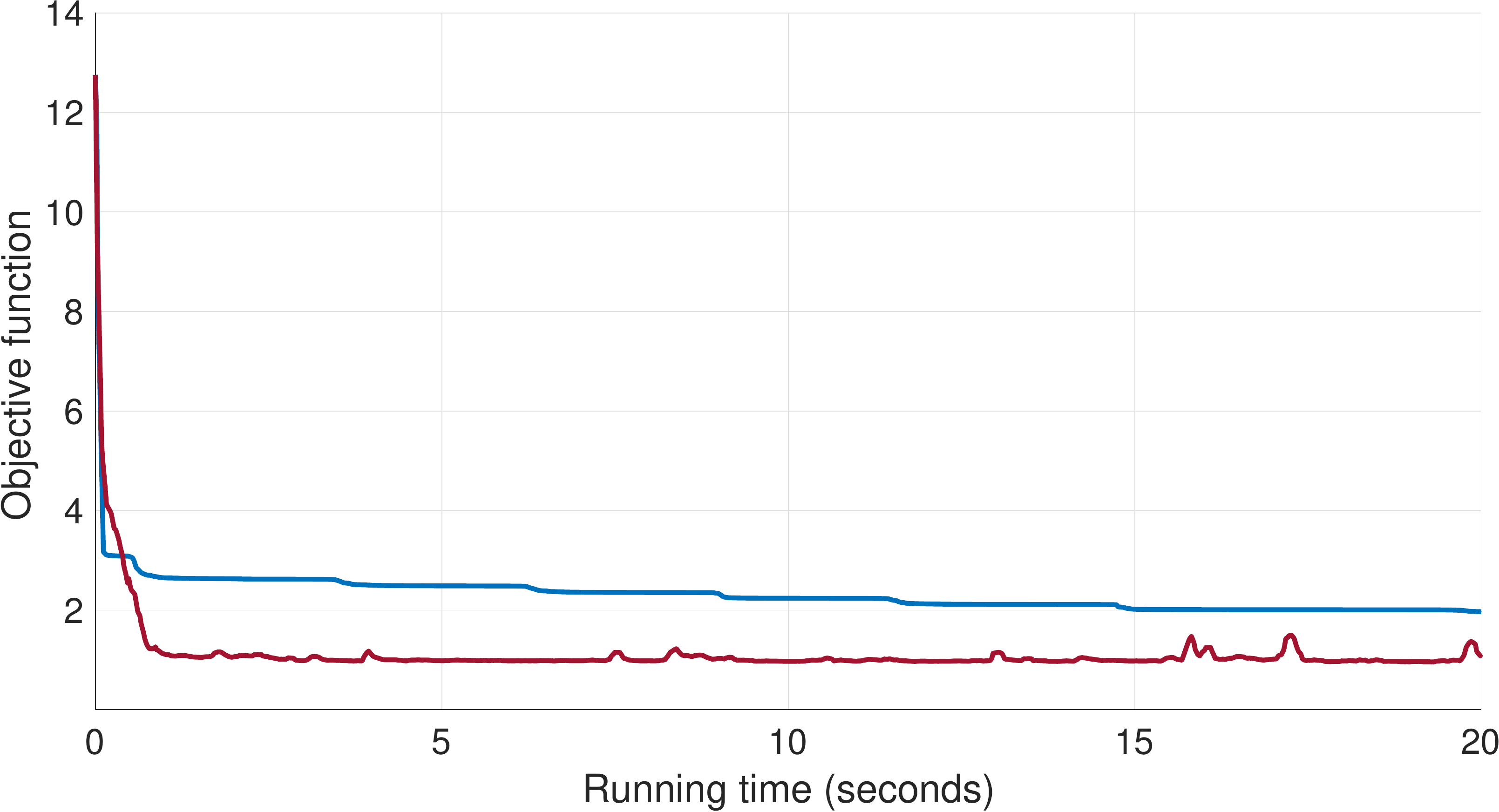}}
	\subfigure[Google]{\label{fig:WeiMniCon_b}\includegraphics[width=0.46\textwidth]{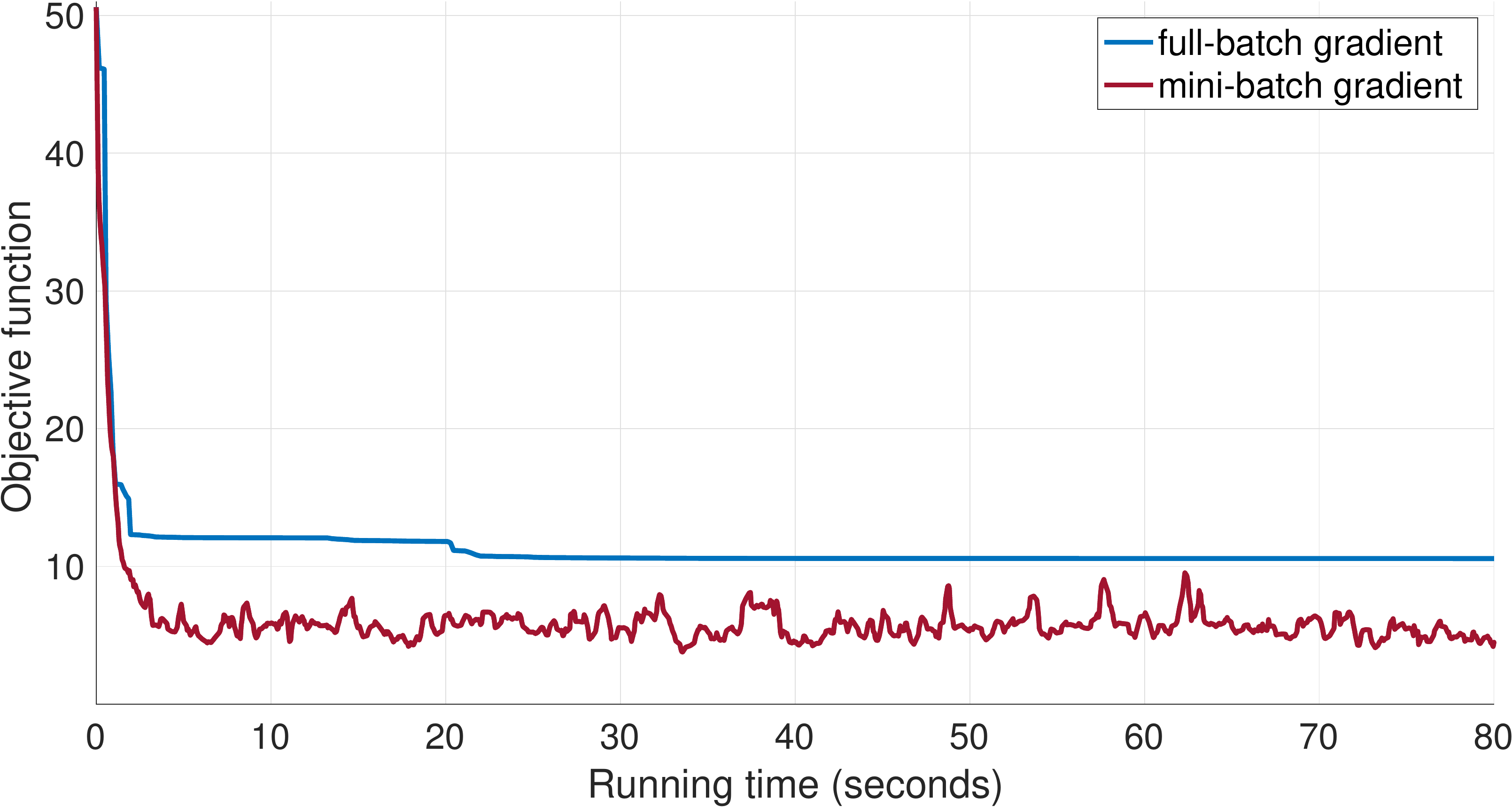}}
	\caption{ Convergence curves of SKCSR (with stochastic gradients estimated from mini-batches $b = 256$) and KCSR (with gradients estimated from full batch (the whole data sequence)) on (a) Weizmann and (b) Google spoken digits datasets. }  
	\label{fig:WeiMniCon}
\end{figure} 

On these datasets, we also observe that evaluation scores of SKCSR are greater than those of KCSR. Thus, we further investigate convergence curves of these models. Figure \ref{fig:WeiMniCon} depicts those of SKCSR and KCSR on Weizmann action videos and Google spoken digits audios, respectively. It is clear that superior performances of SKCSR arise from the exploitation of stochastic gradient descent (SGD) algorithm. SGD allows SKCSR to update its parameter $\bm{\gamma}$ more frequently due to fast estimation of the stochastic gradient. In addition, SGD takes randomness of the data into account and enjoys theoretical guarantee on convergence in an expectation sense \cite{Bottou98}. Therefore, SKCSR is more robust to noise in the data and able to achieve better solution than KCSR.

SKCSR also showed its superior efficiency over the original KCSR and most the other baselines on the ordered MNIST and Acceleration data. Recall that the ordered MNIST data consists of $70K$ samples. Acceleration data contains even much more longer data sequences, where the average length is $125K$. This makes implementation of the memory-demanding methods impossible on regular personal PCs. Among the baselines, only AKS with memory complexity of order $O(r^2)$, where $r \ll n$ is the rank of approximation of the kernel matrix, can handle the ordered MNIST and Acceleration data. However, since AKS employs binary segmentation to sequentially detect the segment boundaries, its solutions are not optimally guaranteed. Visualization of the segmentation results on the Acceleration data in Fig. \ref{fig:AccSeg} and the evaluation scores in the fifth and sixth rows of Table \ref{tab:result_scores} validate the advantages of SKCSR.

\subsubsection{Evaluation of MKCSR}

We next evaluate performance of MKCSR -- an extension of KCSR for handling multiple data sequences. We utilize concatenated Weizmann action videos and MMI Facial AU videos in this experiment. For the Weizmann data, the first, second and third subjects are selected and their corresponding action videos are concatenated to form a long sequence that consists of $30$ segments, each of which belong to one of the ten action categories. For the MMI Facial AU data, videos of all the subjects are concatenated. The new video sequences consists of $491$ frames and $15$ segments. We compare MKCSR with temporal clustering methods, including SSC, TSC and ACA. For all the compared methods, we set the number of clusters $k = 10$ and $k = 15$ for the Weizmann and MMI Facial AU data, respectively, and select the other parameters following the same scheme as mentioned in subsection \ref{sec:exp_params}.

\begin{table}[t]
	\begin{center} 
		\resizebox{0.48\textwidth}{0.03\textheight}{
			\begin{tabular}{|c|c|c|c|c|c|}
				\hline
				\multicolumn{2}{|c|}{Dataset} & SSC & TSC & ACA & MKCSR \\
				\hline \hline
				\multirow{2}{*}{\textbf{Weizmann}} & \mycc ACC & \mycc $0.1496 \; (0.0167)$ & \mycc $0.1967 \; (0.0121)$ & \mycc $0.5743 \; (0.0127)$ & \mycc $\bm{0.8509} \; (0.0139)$ \\
				\cline{2-6}
				& NMI & $0.1439 \; (0.0144)$ & $0.2159 \; (0.0136)$ & $0.5655 \; (0.0106)$ &  $\bm{0.8732} \; (0.0150)$ \\
				\hline
				\multirow{2}{*}{\textbf{MMI Facial AU}} & \mycc ACC & \mycc $0.2315 \; (0.0204)$ & \mycc $0.3890 \; (0.0295)$ & \mycc $0.6757 \; (0.0134)$ & \mycc $\bm{0.9351} \; (0.0103)$ \\
				\cline{2-6}
				& NMI & $0.2432 \; (0.0341)$ & $0.3952 \; (0.0335)$ & $0.6656 \; (0.0198)$ & $\bm{0.9221} \; (0.0095)$ \\
				\hline
			\end{tabular}
		}
	\end{center}
	\caption{Segmentation results on concatenated video sequences from Weizmann and MMI Facial AU datasets returned by different methods. The mean score of each methods over five random runs along with its variance are reported.}
	\label{tab:result_mul}
\end{table}

Fig. \ref{fig:MulSeg} visualizes the segmentation results on multiple video sequences from Weizmann data and Table \ref{tab:result_mul} shows the evaluation scores on both Weizmann and MMI Facial AU datasets. Simultaneous segmentation of multiple data sequences is a challenging task. As we can observe that, in comparison with segmentation results of a single sequence (the second and third rows of Table \ref{tab:result_scores}), evaluation scores of SSC, TSC and ACA on the multiple data sequences are significantly reduced. MKCSR, however, compared to its original method KCSR, could preserve a great amount segmentation accuracy. As we can see that MKCSR obtained up to $0.8509$ of ACC and $0.8732$ of NMI on Weizmann data. For MMI Facial AU data, MKCSR also achieved $0.9351$ of ACC and $0.9221$ of NMI. These results validate that MKCSR can inherit advanced properties from SKCSR to perform efficiently and effectively on multiple data sequences.

\section{Conclusion} \label{sec:conclusion}

Approximation of segmentation for fast computational time and low memory requirement is very important as nowadays more and more large sequential datasets are available. Previous works for approximating optimal segmentation algorithm are either ineffective or inefficient because they still involve in optimization over discrete variables. In this paper, we proposed KCSR to alleviate the aforementioned issue. Our model combines a novel regularization based on sigmoid function with objective of balanced kernel $k-$means to approximate sequence segmentation. Its objective is differentiable almost every where. Therefore, we can use gradient-based algorithm to achieve the optimal segmentation. Note that, our model update all the parameters of interest in an unified manner. This is in contrast to existing approximation methods that sequentially update the segment boundaries, which has no guarantee on quality of the solutions. To further reduce the time and memory complexities, we introduce SKCSR -- a stochastic variant of KCSR. SKCSR employs stochastic gradient descent, where the gradient is estimated from a randomly sampled subsequence, for updating parameters of the model. Thus, it can avoid storing large affinity and/or kernel matrix, which is a critical issue that inhibits existing methods from segmenting long data sequence. Finally, we modify the sigmoid-based regularization to develop MKCSR -- an extended variant of KCSR for simultaneous segmentation of multiple data sequences. Through extensive experiments on various types of sequential data, performances of all the proposed models are evaluated and compared with those of existing methods. The experimental results validates the claimed advantages of the proposed models.

\appendices

\section{Derivation of the gradient} \label{app:grad} 

In this section, we provide derivation of the gradient w.r.t $\bm{\gamma}$. Recall that our objective function is
\begin{equation}
	\begin{split}
		J(\bm{\gamma}) = & \Tr \left( \left( \bm{I}_n - \bm{G}^\top\left(\bm{G}\bm{G}^\top\right)^{-1}\bm{G} \right) \bm{K} \right) \\ 
		& \qquad \qquad \qquad \qquad \qquad + \lambda \Tr(\bm{G} \bm{1} \bm{1}^\top \bm{G}^\top).
	\end{split}    
\end{equation}
The gradient $\nabla \bm{\gamma} = \frac{\partial J}{\partial \bm{\gamma}}$ can be computed using chain rule. We first compute the gradient of $J$ w.r.t $\bm{G}$ as follows:
\begin{equation}
	\begin{split}
		\frac{\partial J}{\partial \bm{G}} = & 2 \left(\bm{G} \bm{G}^\top\right)^{-1} \bm{G} \bm{K} \bm{G}^\top \left(\bm{G} \bm{G}^\top\right)^{-1} \bm{G} \\
		& \qquad \qquad \qquad - 2 \left(\bm{G} \bm{G}^\top\right)^{-1} \bm{G} \bm{K} + \lambda \bm{G} \bm{1} \bm{1}^\top.
	\end{split}
\end{equation}
Since each entry in the $j^{\text{th}}$ column of $\bm{G}$ is a function of continuously segment label $\tau_j$ we need to compute
\begin{equation}
	\frac{\partial G_{i,j}}{\partial \tau_j} = \frac{\partial \text{max}\left( 0, 1 - \lvert \tau_j - i \rvert \right)}{\partial \tau_j} = \begin{cases}
		-1 \quad \text{if } i \leq \tau_j \leq i+1 \\
		\phantom{-}1 \quad \text{if } i-1 \leq \tau_j < i \\
		\phantom{-}0 \quad \text{ otherwise}.
	\end{cases}
\end{equation}
Then the the gradient of $J$ w.r.t $\bm{\tau} = [\tau_1, \hdots, \tau_n]^\top$ is
\begin{equation}
	\frac{\partial J}{\partial \tau_j} = \sum_{i = 1}^{k} \frac{\partial J}{\partial G_{i,j}} \frac{\partial G_{i,j}}{\partial \tau_j} .
\end{equation}
The segment label $\tau_j$ is again computed via a mixture of $k-1$ sigmoid functions, each of whose parameter is $\beta_i$. Thus, we need to compute
\begin{align}
	\frac{\partial \tau_j}{\partial \beta_i} & = \frac{\partial \left(1 + \sum_{i^{\prime}=1}^{k-1} \left(1 + \e^{-\alpha(j-\beta_{i^{\prime}})}\right)^{-1} \right) }{\partial \beta_i} \nonumber \\
	& = - \alpha \left(1 + \e^{-\alpha(j-\beta_i)}\right)^{-1} \left[ 1 - \left(1 + \e^{-\alpha(j-\beta_i)}\right)^{-1}  \right].
\end{align} 
Then the gradient of $J$ w.r.t $\bm{\beta} = [\beta_1, \hdots, \beta_{k-1}]^\top$ can be derived as follows
\begin{equation}
	\frac{\partial J}{\partial \beta_i} = \sum_{j=1}^{n} \frac{\partial J}{\partial \tau_j} \frac{\partial \tau_j}{\partial \beta_i}.
\end{equation}
Finally, we arrive at the gradient of $J$ w.r.t $\bm{\gamma} = [\gamma_1, \hdots, \gamma_k]^\top$
\begin{equation}
	\frac{\partial J}{\partial \gamma_c} = \sum_{i = 1}^{k-1} \frac{\partial J}{\partial \beta_i}  \frac{\partial \beta_i}{\partial \gamma_c}, 
\end{equation}
where
\begin{equation}
	\frac{\partial \beta_i}{\partial \gamma_c} = \begin{cases}
		\frac{(n-1) \e^{\gamma_c}}{\sum_{i^{\prime} = 1}^{k} \e^{\gamma_{i^{\prime}}}} \left( 1 - \frac{ \sum_{i^{\prime} = 1}^{i} \e^{\gamma_{i^{\prime}}} }{ \sum_{i^{\prime} = 1}^{k} \e^{\gamma_{i^{\prime}}} } \right) \quad \text{if } c \leq i, \\
		\phantom{-} -\frac{ (n-1) \e^{\gamma_c}  \sum_{i^{\prime} = 1}^{i} \e^{\gamma_{i^{\prime}}} }{\left( \sum_{i^{\prime} = 1}^{k} \e^{\gamma_{i^{\prime}}} \right)^2} \phantom{--} \; \quad \text{if } c > i.
	\end{cases}
\end{equation}

\newpage


\newpage

\begin{IEEEbiography}[{\includegraphics[width=1in,height=1.25in,clip,keepaspectratio]{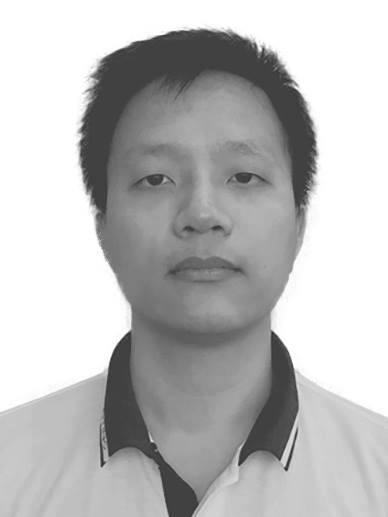}}]{\textbf{Tung Doan}} received the B.S. degrees in Computer Engineering from Hanoi University of Science and Technology in 2014. In 2021, he completed the PhD course at National Institute of Informatics, Japan. He is now a staff lecturer at Department of Computer Engineering, School of Information and Communication Technology, Hanoi University of Science and Technology  His current research interests include deep learning, multiview learning, generative model and sequential data. 
\end{IEEEbiography}

\vspace{-15 cm}

\begin{IEEEbiography}[{\includegraphics[width=1in,height=1.25in,clip,keepaspectratio]{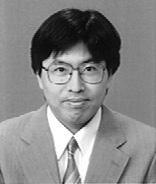}}]{\textbf{Atsuhiro Takasu}} received his B.E., M.E., and Dr.Eng. in 1984, 1986, and 1989, respectively, from the University of Tokyo, Japan. He is a professor at the National Institute of Informatics, Japan. His research interests are data engineering and data mining. He is a member of the ACM, IEEE, IEICE, IPSJ, and JSAI.
\end{IEEEbiography}

%

\EOD

\end{document}